\documentclass[letterpaper]{article} 
\usepackage{aaai2026}  
\usepackage{times}  
\usepackage{helvet}  
\usepackage{courier}  
\usepackage[hyphens]{url}  
\usepackage{graphicx} 
\usepackage{natbib}  
\usepackage{caption} 
\usepackage{algorithm}
\usepackage[noend]{algorithmic}
\usepackage{adjustbox}
\usepackage{tikz}
\usepackage{multicol}
\usepackage{multirow}
\usepackage{booktabs}
\usepackage{amsmath,amsfonts}
\usepackage{amsthm}
\usepackage{xcolor}
\usepackage{colortbl}
\usepackage{threeparttable}
\usepackage{bbding}
\usepackage{subcaption}
\usepackage{enumitem}
\usepackage{newfloat}
\usepackage{listings}
\DeclareCaptionStyle{ruled}{labelfont=normalfont,labelsep=colon,strut=off} 
\floatstyle{ruled}
\newfloat{listing}{tb}{lst}{}
\floatname{listing}{Listing}
\title{DeNAS-ViT: Data Efficient NAS-Optimized Vision Transformer\\for Ultrasound Image Segmentation}
\author{
    Renqi Chen\textsuperscript{\rm 1},
    Xinzhe Zheng\textsuperscript{\rm 2},
    Haoyang Su\textsuperscript{\rm 3}\thanks{Corresponding author.},
    Kehan Wu\textsuperscript{\rm 4}
}
\affiliations{
    \textsuperscript{\rm 1}College of Intelligent Robotics and Advanced Manufacturing, Fudan University, Shanghai, China\\
    \textsuperscript{\rm 2}School of Computing, National University of Singapore, Singapore\\
    \textsuperscript{\rm 3}Australian Institute for Machine Learning, The University of Adelaide, Adelaide, Australia\\
    \textsuperscript{\rm 4}Southern University of Science and Technology, Shenzhen, China\\
    
    rqchen23@m.fudan.edu.cn, zxz.krypton@outlook.com, haoyang.su@adelaide.edu.au, 12010519@mail.sustech.edu.cn

}

\usepackage{bibentry}

\definecolor{iccvblue}{rgb}{0.21,0.49,0.74}

\begin{document}

\maketitle

\begin{abstract}
Accurate segmentation of ultrasound images is essential for reliable medical diagnoses but is challenged by poor image quality and scarce labeled data. Prior approaches have relied on manually designed, complex network architectures to improve multi-scale feature extraction. However, such handcrafted models offer limited gains when prior knowledge is inadequate and are prone to overfitting on small datasets. In this paper, we introduce DeNAS-ViT, a \textbf{D}ata \textbf{e}fficient \textbf{NAS}-optimized \textbf{Vi}sion \textbf{T}ransformer, the first method to leverage neural architecture search (NAS) for ultrasound image segmentation by automatically optimizing model architecture through token-level search. Specifically, we propose an efficient NAS module that performs multi-scale token search prior to the ViT’s attention mechanism, effectively capturing both contextual and local features while minimizing computational costs. Given ultrasound’s data scarcity and NAS’s inherent data demands, we further develop a NAS-guided semi-supervised learning (SSL) framework. This approach integrates network independence and contrastive learning within a stage-wise optimization strategy, significantly enhancing model robustness under limited-data conditions. Extensive experiments on public datasets demonstrate that DeNAS-ViT achieves state-of-the-art performance, maintaining robustness with minimal labeled data. Moreover, we highlight DeNAS-ViT’s generalization potential beyond ultrasound imaging, underscoring its broader applicability.
\end{abstract}


\section{Introduction}

Ultrasound is a critical medical imaging tool for diagnosing cardiac diseases, and precise segmentation of ultrasound images can greatly aid clinicians in thoroughly analyzing cardiac conditions~\cite{leclerc2019deep}. 
Nonetheless, the development of robust and efficient segmentation algorithms faces challenges such as the scarcity of annotated data, speckle noise in ultrasound images, and the difficulty of distinguishing adjacent anatomical structures~\cite{zhou2023dsanet,lin2023samus}. These challenges highlight the need to \textbf{enhance feature extraction across multiple scales}~\citep{lin2017feature,su2020adapting,chen2023sshnn} and to \textbf{improve model robustness to limited data}~\citep{tarvainen2017mean,qiao2018deep,xia2020uncertainty}.

\begin{figure}[t]
    \centering
    \includegraphics[width=0.99\linewidth]{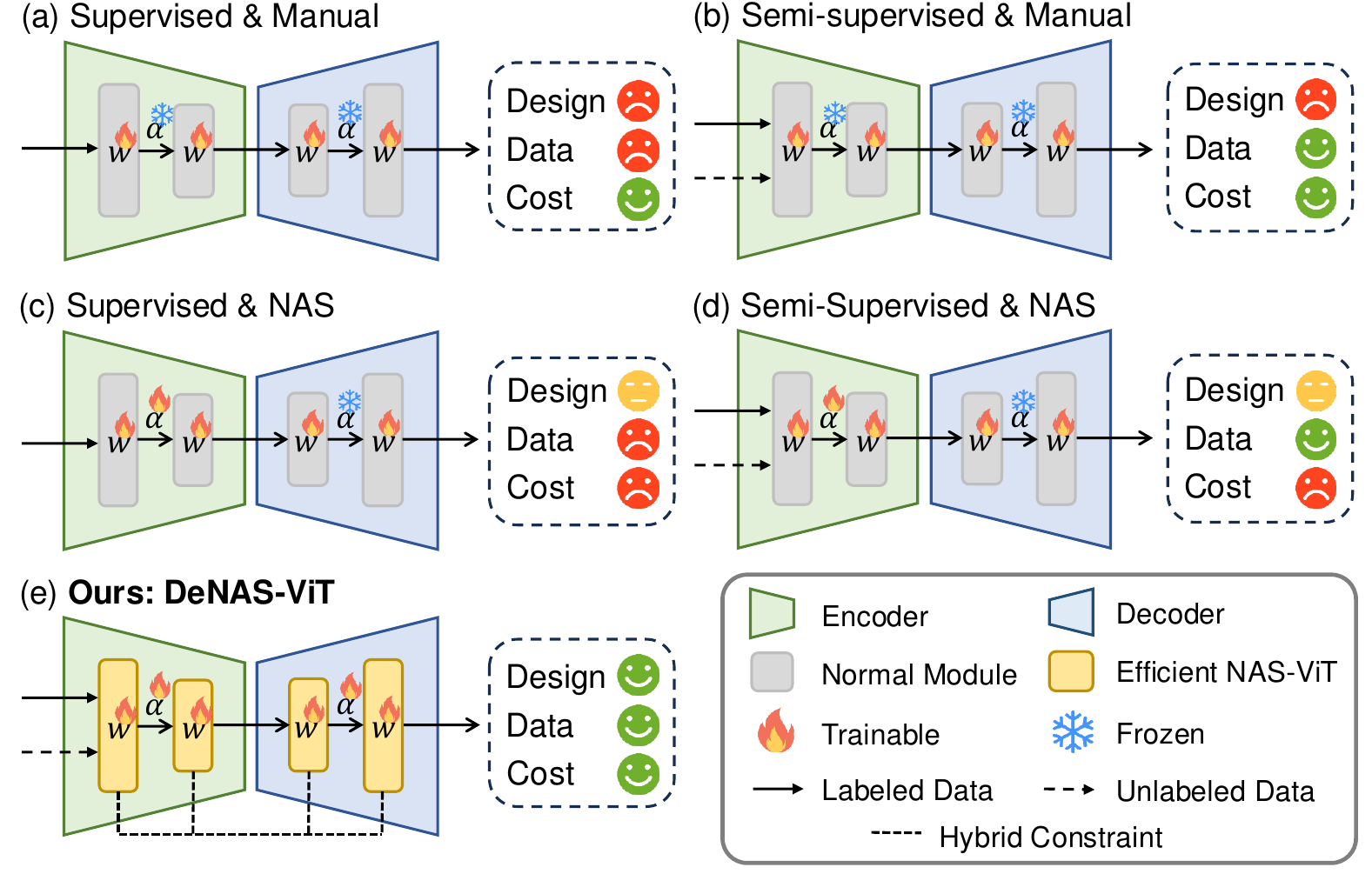}
    \vspace{-2mm}
    \caption{An illustration comparing DeNAS-ViT with existing baselines for segmentation task. Use smiley, neutral, and sad faces to present the performance. ``Design'' denotes the optimization of model architecture for multi-scale feature extraction, ``Data'' denotes the robustness to limited data, and ``Cost'' denotes the computational resource consumption. $\alpha$ and $w$ present architecture and weights, respectively.}\label{first_figure}
\end{figure}

In recent years, deep learning has advanced medical image segmentation by improving multi-scale feature understanding. Ronneberger et al.~\cite{ronneberger2015u} introduced U-Net, a U-shaped encoder-decoder architecture with exceptional feature extraction capabilities, inspiring variants like UNet++~\cite{zhou2019unet++} that emphasize local feature perception. Concurrently, the Vision Transformer (ViT)~\cite{dosovitskiy2020image} emerged, leveraging self-attention to capture long-range pixel relationships and contextual information. This led to hybrid models such as TransUNet~\cite{chen2021transunet}, which integrates U-Net with ViT to balance local and global features, and EfficientViT~\cite{cai2023efficientvit}, which employs multi-scale linear attention for efficient feature representation. However, manually designing such architectures demands significant expertise and struggles to optimize multi-scale feature extraction effectively.
NAS has gained traction as an automated solution, optimizing network structures within a defined search space~\cite{liu2019auto,xu2019pc,lu2022m}. While NAS has been applied to ultrasound image segmentation~\cite{cao2022auto,qian2022hasa,chen2023sshnn}, existing methods typically select operations at the module level (e.g., convolutions or Transformer blocks), neglecting finer-grained operations within these modules. This limitation restricts the representational capacity for multi-scale feature extraction. Additionally, module-level search strategies often suffer from low precision, resulting in inefficient computational resource utilization. In this work, we address both challenges by refining operation selection within modules.

Moreover, ultrasound datasets often lack sufficient labeled data, a challenge exacerbated by NAS’s data-intensive nature. SSL offers a solution by leveraging unlabeled data, with methods like mean teacher~\cite{tarvainen2017mean,yu2019uncertainty} and co-training~\cite{chen2021semi,miao2023caussl} enhancing robustness through constraints on unlabeled data. Combining NAS with SSL thus holds promise for addressing both limited labeled data and complex model design in ultrasound segmentation. However, prior efforts~\cite{pauletto2022se,chen2023sshnn} have largely adopted basic NAS-SSL integration without embedding additional constraints during NAS training, often leading to overfitting on small labeled datasets, especially with complex architectures~\cite{huesmann2021impact,song2023efficient,salehin2024automl}.

To address these challenges, we propose DeNAS-ViT, a data-efficient, NAS-optimized ViT designed for ultrasound image segmentation. Fig.~\ref{first_figure} highlights the key distinctions from existing methods. For architecture design, we employ NAS at multiple levels: 
at the cellular level, we introduce an efficient NAS-ViT module that integrates NAS with ViT to optimize multi-scale token representations while reducing computational overhead; at the module level, NAS is applied distinctly to the encoder and decoder, each with specialized search spaces to preserve their unique roles. 
To address ultrasound’s data scarcity and NAS’s data-intensive nature, we propose a NAS-based SSL framework with hybrid constraints, which includes: (1) a NAS-derived network independence loss to encourage complementary model representations; (2) a hierarchical NAS-based contrastive loss to maximize mutual information across views, boosting feature representation and generalization; and (3) a tailored stage-wise optimization strategy. In summary, the main contributions of this paper are:

\begin{itemize}
\item We propose a data efficient NAS-optimized Vision Transformer (DeNAS-ViT) for ultrasound segmentation, marking the first integration of NAS for token-level searching and multi-scale feature representation.

\item We introduce a NAS-based constraint-driven SSL framework with a multi-stage optimization strategy, reducing overfitting with limited ultrasound image annotations and alleviating the data-intensive demands of NAS.

\item Experiments on public datasets show that our method achieves state-of-the-art performance and shows potential for generalization beyond ultrasound segmentation.
\end{itemize}

\section{Related Work}
\noindent \textbf{Neural Architecture Search.}~
NAS is proposed to address the difficulty of network design and find the optimal network architecture within the search space~\cite{ren2021comprehensive,white2023neural}. Differentiable architecture search (DARTS)~\cite{liu2018darts} created a continuous relaxation algorithm to make gradient-based NAS efficiently trainable. Then, PC-DARTS~\cite{xu2019pc} proposed partial channel connections and edge normalization, reducing GPU memory consumption. Edge normalization inspired the hierarchical NAS (HNAS)~\cite{liu2019auto,yu2023hct,yang2023dast}, enabling a multi-level architecture search. The Transformer is also used for context information. \citet{yang2023dast} proposed DAST which incorporates the ViT layer as candidate operations for cell-level searching. Unlike these approaches, we delve into a more fine-grained search space to enhance multi-scale feature representation and reduce parameter costs.

\noindent \textbf{SSL for Medical Image Segmentation.}~
SSL addresses the scarcity of labeled data in medical image segmentation, with approaches broadly classified into pseudo-labeling~\cite{tarvainen2017mean,bai2017semi,wang2021semi} and consistency regularization~\cite{qiao2018deep,yu2019uncertainty,xia2020uncertainty}. Pseudo-labeling methods generate labels for unlabeled data during training, as seen in self-training~\cite{bai2017semi}. In contrast, consistency regularization enforces prediction alignment. Tarvainen et al.~\cite{tarvainen2017mean} proposed the Mean Teacher (MT), averaging weights to create a teacher model that ensures consistency between predictions and targets. Beyond MT, Qiao et al.~\cite{qiao2018deep} developed a deep co-training framework, training two networks on distinct views for complementary learning. In this work, we propose a hybrid constraint-driven SSL framework that not only mitigates the limited availability of medical image data but also reduces the sample dependency of NAS.

\begin{figure}[t]
    \centering
    \includegraphics[width=0.99\linewidth]{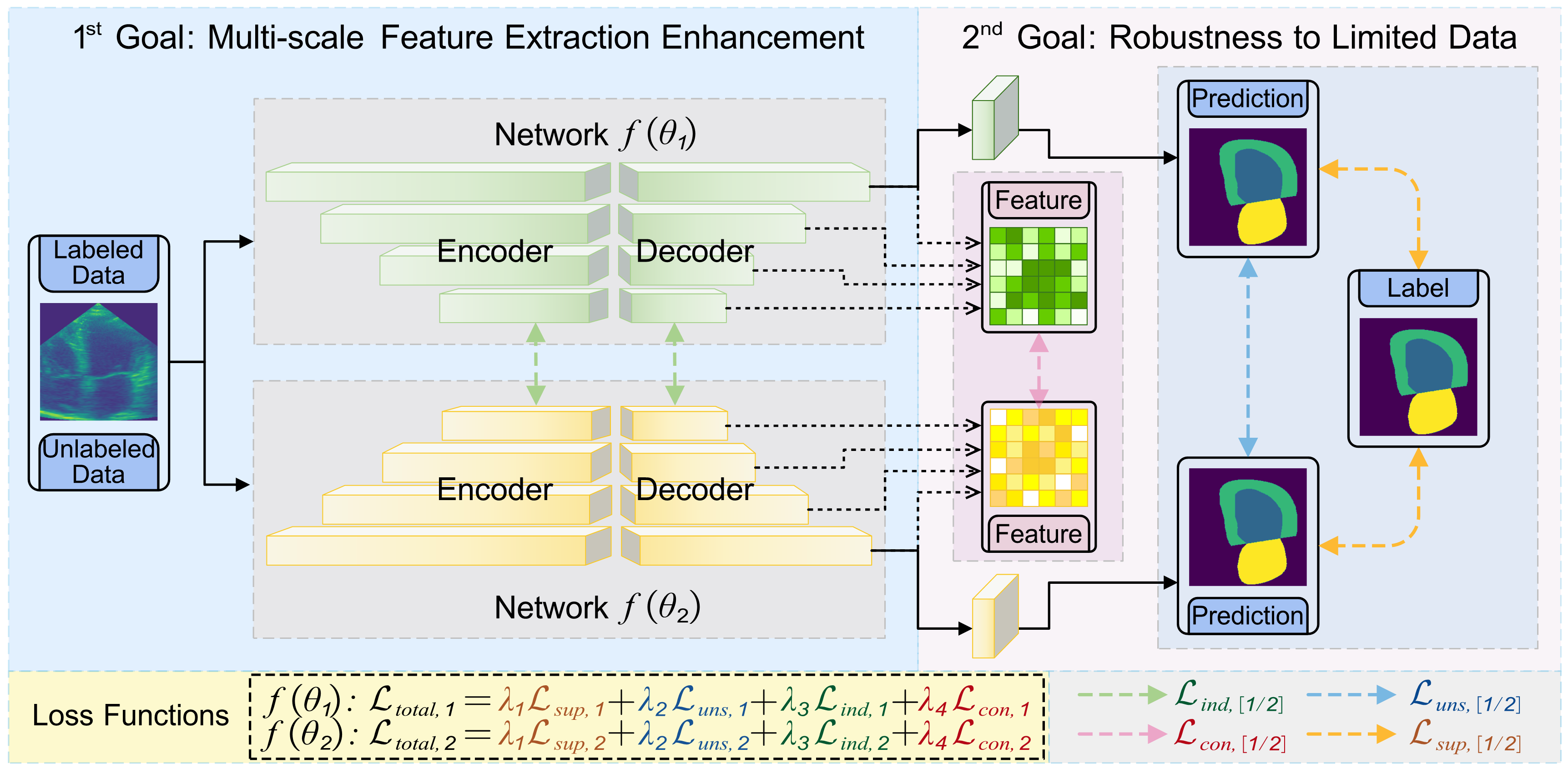}
    \vspace{-1mm}
    \caption{The pipeline of DeNAS-ViT. The hierarchical structure of NAS networks serves for the goal of multi-scale feature extraction enhancement, and multi-constraint SSL serves for the goal of robustness to limited data.}\label{Overview}
\end{figure}

\begin{figure*}[t]
    \centering
    \includegraphics[width=0.96\linewidth]{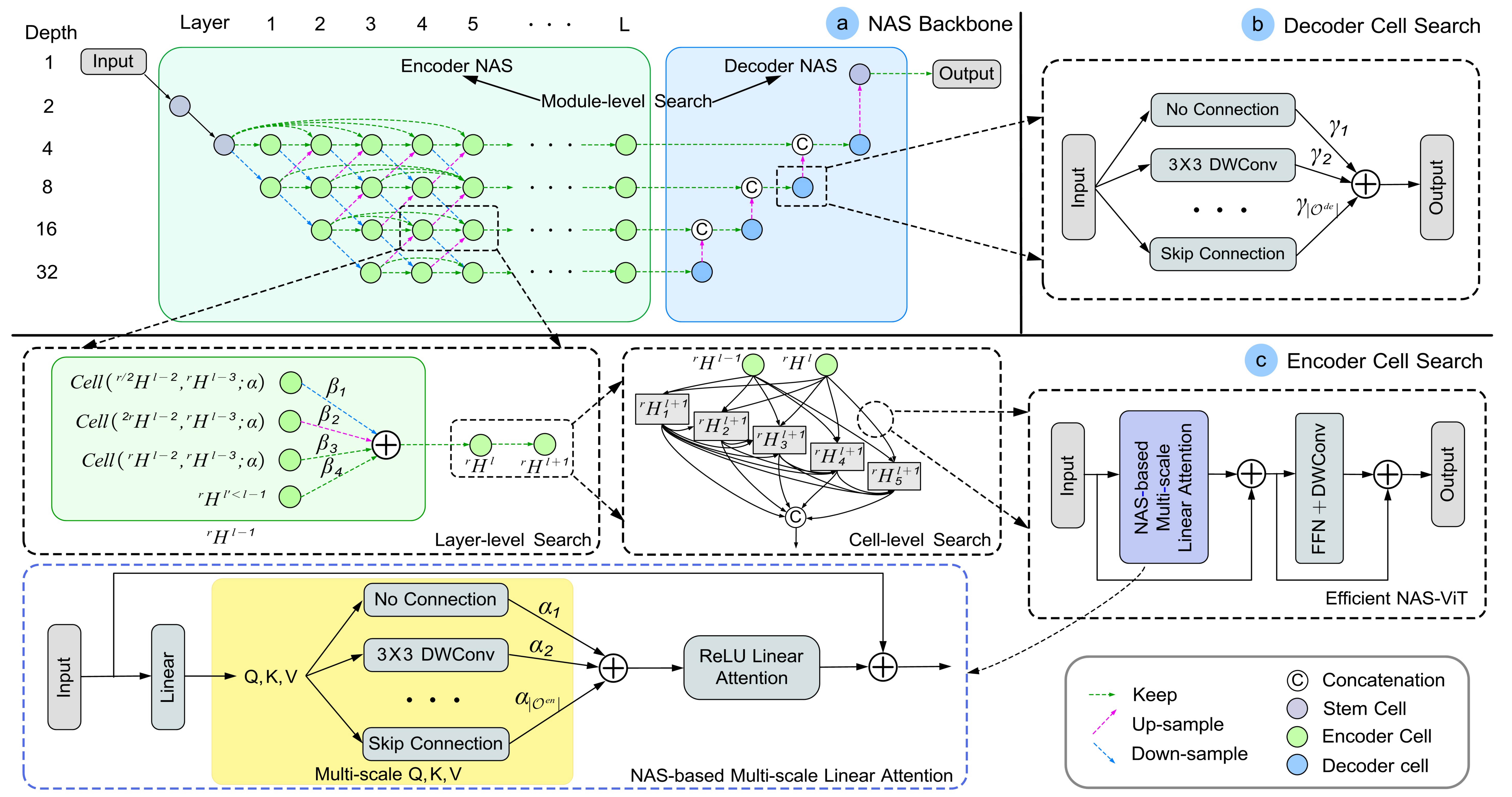}
    \vspace{-2mm}
    \caption{The proposed NAS backbone. (a) shows an overview of the NAS backbone, which consists of an encoder NAS and a decoder NAS, representing a module-level search. The input is passed through the encoder NAS to obtain multi-resolution feature maps, where a hierarchical encoder cell search is performed (shown in (c)). These outcomes are then processed by the decoder cells (shown in (b)), which concatenate features and complete the recovery process through further searching.}\label{Structure}
\end{figure*}

\section{Methodology}
An overview of DeNAS-ViT is presented in Fig.~\ref{Overview}. To enhance multi-scale feature extraction, we introduce the Efficient NAS-ViT module (Sec.\ref{efficientvit}) within the NAS backbone (Sec.~\ref{NAS backbone}). This module integrates NAS with a ViT to perform token-level search, optimizing multi-scale feature representation while alleviating the computational burden typically associated with NAS.

To improve robustness under limited labeled data, we propose a hybrid constraint-driven semi-supervised learning (SSL) framework (Sec.\ref{SSL loss}), extending a co-training approach. This framework employs two networks, $f(\theta_1)$ and $f(\theta_2)$, which share the NAS backbone but operate with distinct parameters and independent optimizers. We implement a stage-wise optimization strategy (Sec.~\ref{optimization_strategy}), incorporating an independence loss to promote complementary representations between the networks, alongside a contrastive loss for enhanced feature representation. This dual approach refines model updates, improving resilience to data scarcity.

\vspace{-1mm}
\subsection{Efficient NAS-ViT Module}\label{efficientvit}

In DAST~\cite{yang2023dast}, the ViT layer is treated as the candidate operation for multi-scale feature representation, leading to increased computation complexity. On the other hand, since the ViT is more complex than other operations, the model tends to select the ViT during backbone optimization, rendering the search meaningless. To solve these, we use the ViT as the basic unit, further implementing NAS before the attention calculation for searching multi-scale tokens, termed Efficient NAS-ViT. Compared to EfficientViT \cite{cai2023efficientvit}, where tokens are processed by fixed multi-scale convolution and sent into attention separately, our design can obtain a better multi-scale feature representation while reducing excessive parameter overhead. The effectiveness of this novel Transformer-based approach for NAS is proven through the empirical results (Sec. \ref{ablation}). An overview of the Efficient NAS-ViT is shown in the ``Efficient NAS-ViT'' of Fig. \ref{Structure}(c), used in the encoder NAS's cell search. 

Given an input $x$, after a linear projection, the tokens are defined as $Q=xW_{Q}$, $K=xW_{K}$, and $V=xW_{V}$, where $W_{Q}$, $W_{K}$, and $W_{V}$ represent the learnable linear projection matrices.
Rather than using fixed operations to obtain multi-scale tokens, which constrains the token representation and affects attention calculation, we employ NAS to search for token representations, thereby enhancing the multi-scale learning capability. After incorporating partial channel connections to reduce memory overhead and reusing continuous relaxation for a differentiable search space, the multi-scale tokens $Q'/K'/V'$ are:
\begin{equation}
\begin{aligned}
    \{Q'/K'/&V'\} = (1-\textbf{P})\odot \{Q/K/V\}+\\ &\sum_{O_{i}\in \mathcal{O}^{en}}\frac{exp\{\alpha_{i}\}}{\sum_{j=1}^{|\mathcal{O}^{en}|}exp\{\alpha_{j}\}}\cdot O_{i}(\textbf{P}\odot\{Q/K/V\}),
\end{aligned}
\end{equation}
where $\textbf{P}$ is the sampling mask for channel selection, $\odot$ denotes Hadamard product, $O_{i}$ denotes the $i$-th operation selected from the set of encoder candidate operations $\mathcal{O}^{en}$ (encoder and decoder consider different sets of candidate operations due to their respective roles, please refer to Appendix A), and $\alpha$ is the cell architecture parameter, which measures the weight of the related candidate operation. Subsequently, tokens are computed by ReLU linear attention for contextual information, denoted as:
\begin{align}
  R_{n}=\sum_{i=1}^{S}\frac{ReLU(Q'_{n})ReLU(K'_{i})^{T}}{\sum_{j=1}^{S}ReLU(Q'_{n})ReLU(K'_{j})^{T}}V'_{i},
\end{align}
where subscript $n$ denotes the $n$-th row of $R/Q$, and $S$ is the sequence length. Then, the output $R$ is fed into the FFN+DWConv layer (the application of depthwise convolution on the FFN layer) for local information capture. 

\subsection{NAS Backbone}\label{NAS backbone}
Our NAS backbone builds on HNAS framework~\citep{liu2019auto,fang2020densely,chen2023sshnn}, with enhancements for multi-scale optimization: (1) macro-level dual decoder NAS; (2) micro-level efficient NAS-ViT.
\subsubsection{Encoder NAS} We have cell- and layer-level searches for the multi-scale feature extraction. There are $N$ intermediate nodes in cell search space, where the edge between nodes corresponds to the proposed Efficient NAS-ViT module, shown in the ``Cell-level Search'' of Fig.~\ref{Structure}(c). After processing inputs by Efficient NAS-ViT ($f_{NAS-ViT}$), the output of the $n$-th node of the encoder cell in the $l$-th layer is $
    ^{r}H_{n}^{l} = \sum_{^{r}H_{i}^{l}\in\mathcal{I}}f_{NAS-ViT}(^{r}H_{i}^{l};\alpha)$, 
where the input set $\mathcal{I}$ includes the previous cell's output and previous nodes' outputs in the current cell, and $r$ denotes resolution. 

To capture multi-resolution features, our model considers six values: $r=1, 2, 4, 8, 16, 32$, where $r=1$ corresponds to the original image size. Two stem cells down-sample the input from $r=1$ to $r=4$.
When $r=4$, the spatial size of the feature maps is $1/4$ of the case when $r=1$.
The final output tensor of the cell is the concatenation of outputs from all nodes. For simplicity, the cell-level search is written as $Cell_{\alpha}(^{r}H^{l-1},^{r}H^{l-2})$.

Following~\cite{fang2020densely}, the layer-level search aims at aggregating feature paths from different resolutions by employing relaxation and skip connections, shown in the ``Layer-level Search'' of Fig.~\ref{Structure}(c). Path weights are referred to as architecture parameters $\beta$. There are $L$ layers in the backbone, and the $l$-th layer level search is denoted as:
\begin{equation}
\begin{aligned}
  ^{r}H^{l} & = \beta_{1}Cell(^{\frac{r}{2}}H^{l-1},^{r}H^{l-2})+\beta_{2}Cell(^{2r}H^{l-1},^{r}H^{l-2})\\
  &+\beta_{3}Cell(^{r}H^{l-1},^{r}H^{l-2})+\beta_{4}\{^{r}H^{l'}\in{^{r}H}|l'<l-1\},
\end{aligned}
\end{equation}
where $\sum_{i}\beta_{i}=1$, normalized and implemented as softmax.
\subsubsection{Decoder NAS}
Rather than employing fixed convolution kernel sizes, we also introduce NAS for the decoder to enhance the capability. A U-shaped architecture is used as the backbone for the decoder NAS, as depicted in Fig. \ref{Structure}(a).  

In the decoder NAS, the features from the $r=32$ layer after the encoder NAS are fed into the decoder cell as the initial input. To mitigate the high computational complexity introduced by NAS, the decoder cell does not contain intermediate nodes, as shown in Fig. \ref{Structure}(b). The search process can be represented as:
\begin{align}
    ^{r}H^{de} = \sum_{O_{i}\in\mathcal{O}^{de}}\frac{exp\{\gamma_{i}\}}{\sum_{j=1}^{|\mathcal{O}^{de}|}exp\{\gamma_{j}\}}\cdot O_{i}(^{r}H^{l=L}),
    \end{align}
where $\gamma$ are decoder architecture parameters, and $^{r}H^{de}$ is the final output tensor of the decoder cell in $r$-resolution.

After upsampling, the features $^{r}H^{de}$ are concatenated with $^{r/2}H^{L}$, and a convolution layer is applied to match the number of channels with the $r/2$ decoder NAS. This process is repeated until the $r=4$ features are combined. Finally, an upsample layer is employed to recover the outputs to the full resolution, followed by a convolution layer to obtain the desired number of classes for target tasks.
\subsection{NAS-based Constraint-driven SSL}\label{SSL loss}
To leverage the unlabeled data, the co-training framework of SSL is implemented, where the unsupervised loss is:
\begin{align}\label{formula_uns}
\mathcal{L}_{uns,a}=CE(p_{u,a},\hat{y}_{u,b}),
\end{align}
where $a,b\in\{1,2\}$ and $a\neq b$, corresponds to $f(\theta_{a})$ and $f(\theta_{b})$, respectively. $CE(\cdot)$ indicates the cross-entropy loss. $p_{u,a}$ is the predicted probability map generated by one network on the unlabeled data, and $\hat{y}_{u,b}$ is the corresponding one-hot pseudo label generated by another.

As the algorithmic independence could facilitate the creation of distinct networks~\cite{miao2023caussl}, especially in the co-training, complementary networks are capable of capturing diverse feature information and are less likely to overfit to a particular subset. Thus, we incorporate network independence loss and propose a stage-wise optimization strategy (Sec. \ref{optimization_strategy}) to fully utilize very few samples. The network independence loss is defined between two NAS backbones based on convolutional layers at the same position:
\begin{equation}\label{formula:ind}
  \begin{aligned}
    \mathcal{L}_{ind,a}=\frac{1}{L_{CNN}}\sum_{i=1}^{L_{CNN}}IND(\theta_{a,i},\theta_{b,i};G_{b,i}),
\end{aligned}  
\end{equation}
where $L_{CNN}$ is the number of convolutional layers. $\theta_{a,i}\in\mathbb{R}^{C_{out}\times d}(d=K\times K\times C_{in})$ are the weights of the $i$-th convolutional layer in $f(\theta_{a})$, reshaped into a matrix form, where $K$ is kernel size. $C_{in}$, $C_{out}$ are the number of input and output feature channels. $G_{b,i}\in\mathbb{R}^{C_{out}\times C_{out}}$ is the corresponding optimal coefficient matrix. Given matrices $A,B, G_{B}$, the independence loss function is defined as:
$
    IND(A,B;G_{B})=\frac{1}{C_{out}}\sum_{i=1}^{C_{out}}(\frac{\boldsymbol{v}_{A,i}\cdot\boldsymbol{q}_{B,i}}{|\boldsymbol{v}_{A,i}|\times|\boldsymbol{q}_{B,i}|})^{2}
$,
where $\boldsymbol{v}_{A,i}$ is the $i$-th row of $A$ and $\boldsymbol{q}_{B,i}=(G_{B}\times B)_{i}$. 

Since algorithmic independence essentially endows the network with the capability to observe the same image from different perspectives, we consider that contrastive loss can be utilized to maximize the mutual information across these views to enhance discriminative feature representation, which also helps the model to generalize to new samples. Benefiting from network independence, there is also no need to construct an asymmetric architecture for contrastive loss. In this work, an uncertainty-based contrastive loss~\cite{huang2023semi} is calculated based on the hierarchical architecture of NAS, measured at different stages of the decoder NAS cell between two networks. The uncertainty estimation is defined using smoothed KL-divergence and considers features at various resolutions:
\begin{equation}
    U_{a}^{r,h,w}=\sum_{c=0}^{C-1}(^{r}_{a}H^{de})^{c,h,w}\cdot\log\frac{(^{r}_{a}H^{de})^{c,h,w}+\epsilon}{\overline{(^{r}_{a}H^{de})^{c}}+\epsilon},
\end{equation}
where $C$ is the channel dimension, $\epsilon$ is a small bias term, $r\in\{4,8,16,32\}$ is used, and $^{r}_{a}H^{de}$ is the output tensor of the decoder cell at the $r$-resolution. $\overline{(\cdot)}$ is the mean value across the channel dimension. A higher estimation value reflects a higher uncertainty, which can be used to compel the lower-quality features to align with their higher-quality counterparts~\cite{huang2023semi}. The positions of these features are estimated by $\mathcal{P}_{a}^{r,h,w}=\textbf{1}\odot \{U_{a}^{r,h,w}>U_{b}^{r,h,w}\}$.
Then, mean squared error is used in the loss function:
\begin{equation}\label{formula:con}
\begin{aligned}
    \mathcal{L}_{con,a}=\sum_{r\in\{4,8,16,32\}}\sum_{p\in\mathcal{P}_{a}^{r,h,w}}MSE((^{r}_{a}H^{de})^{p},(^{r}_{b}H^{de})^{p}).
\end{aligned}
\end{equation}

\begin{table*}[t]
  \centering
   \begin{adjustbox}{width=0.99\linewidth} 
\begin{tabular}{c|ccc|ccc|ccc}
    \toprule  
    \multirow{2}{*}{Method}& \multicolumn{3}{c|}{HMC-QU}& \multicolumn{3}{c|}{CAMUS} & \multicolumn{3}{c}{CETUS} \\
    \cmidrule(){2-10} 
     &DSC$\uparrow$ & IoU$\uparrow$ & 95HD$\downarrow$ & DSC$\uparrow$& IoU$\uparrow$ &95HD$\downarrow$ & DSC$\uparrow$& IoU$\uparrow$ &95HD$\downarrow$ \\
    \midrule  
     UNet++  & $0.899_{(0.005)}*$ & $0.898_{(0.004)}*$ & $4.860_{(0.102)}*$ & $0.919_{(0.006)}*$ & $0.855_{(0.009)}*$ & $6.584_{(0.578)}*$ &  $0.952_{(0.002)}*$ & $0.968_{(0.001)}*$ & $2.386_{(0.091)}*$  \\
    
     nnU-Net  & $0.908_{(0.005)}*$ & $0.907_{(0.005)}*$ & $3.843_{(0.310)}*$ & $0.922_{(0.003)}*$ & $0.860_{(0.004)}*$ & $6.075_{(0.412)}*$ & \underline{$0.958_{(0.001)}*$} & $0.970_{(0.001)}*$ & \underline{$2.099_{(0.048)}*$}  \\
    
     Transfuse  & $0.903_{(0.004)}*$ & $0.903_{(0.004)}*$ & $4.304_{(0.228)}*$ & \underline{$0.923_{(0.004)}*$} & \underline{$0.861_{(0.006)}*$} & \underline{$5.853_{(0.496)}$} & $0.957_{(0.002)}*$ & \underline{$0.971_{(0.001)}*$} & $2.152_{(0.102)}*$  \\
    
    \cmidrule(){1-10}
     Auto-DeepLab &$0.908_{(0.004)}*$ & $0.907_{(0.004)}*$ & $3.857_{(0.164)}*$ & $0.918_{(0.003)}*$ & $0.851_{(0.006)}*$ & $6.641_{(0.473)}*$ & $0.954_{(0.004)}*$ & $0.968_{(0.001)}*$ & $2.278_{(0.133)}*$  \\
    
    M\textsuperscript{3}NAS  & $0.910_{(0.007)}*$ & $0.909_{(0.006)}*$ & \underline{$3.709_{(0.288)}*$} & $0.920_{(0.004)}*$ & $0.856_{(0.004)}*$ & $6.405_{(0.556)}*$ & $0.956_{(0.005)}*$ & $0.969_{(0.004)}$ & $2.175_{(0.174)}*$  \\
    
     GeNAS   & \underline{$0.913_{(0.004)}*$} & $0.908_{(0.005)}*$ & $3.748_{(0.219)}*$ & $0.917_{(0.003)}*$ & $0.852_{(0.003)}*$ & $6.782_{(0.425)}*$  & $0.949_{(0.002)}*$ & $0.966_{(0.001)}*$ & $3.267_{(0.119)}*$ \\
    \midrule
     URPC$^{\dagger}$   & $0.892_{(0.004)}*$ & $0.891_{(0.004)}*$ & $5.355_{(0.125)}*$ & $0.912_{(0.002)}*$ & $0.842_{(0.002)}*$ & $6.708_{(0.152)}*$ & $0.941_{(0.004)}*$ & $0.963_{(0.003)}*$ & $3.374_{(0.185)}*$  \\
    
    CPS$^{\dagger}$  & $0.878_{(0.005)}*$ & $0.877_{(0.005)}*$ & $6.908_{(0.296)}*$ & $0.901_{(0.004)}*$ & $0.827_{(0.005)}*$ & $8.864_{(0.477)}*$  & $0.923_{(0.005)}*$ & $0.954_{(0.003)}*$ & $4.306_{(0.282)}*$ \\
    
    CnT-B$^{\dagger}$  & {$0.911_{(0.006)}*$} & \underline{$0.908_{(0.005)}*$} & $4.102_{(0.496)}*$
    & $0.917_{(0.003)}*$ & $0.847_{(0.002)}*$ & $6.846_{(0.421)}*$ & $0.943_{(0.004)}*$ & $0.959_{(0.003)}*$ & $3.488_{(0.341)}*$  \\
    
    ARCO-SG$^{\dagger}$ & $0.908_{(0.003)}*$ & $0.905_{(0.004)}*$ & $3.912_{(0.203)}*$ & $0.916_{(0.004)}*$ & $0.850_{(0.003)}*$ & $6.410_{(0.494)}*$ & $0.948_{(0.005)}*$ & $0.964_{(0.004)}*$ & $3.125_{(0.282)}*$  \\
    
    \cmidrule(){1-10}
    Se\textsuperscript{2}NAS$^{\dagger}$   & $0.907_{(0.004)}*$ & $0.907_{(0.004)}*$ & $3.941_{(0.303)}*$ & $0.920_{(0.002)}*$ & $0.856_{(0.003)}*$ & $6.410_{(0.311)}*$ & $0.955_{(0.003)}*$ & $0.967_{(0.003)}*$ & $2.235_{(0.111)}*$  \\
    
    SSHNN$^{\dagger}$  & $0.906_{(0.002)}*$ & $0.904_{(0.002)}*$ & $4.011_{(0.145)}*$
    & $0.922_{(0.002)}*$ & $0.859_{(0.003)}*$ & $6.116_{(0.278)}*$ & $0.949_{(0.002)}*$ & $0.961_{(0.001)}*$ & $3.053_{(0.087)}*$  \\

    \cellcolor[gray]{0.9}{\textbf{DeNAS-ViT}$^{\dagger}$}  & \cellcolor[gray]{0.9}{{$\boldsymbol{0.933_{(0.002)}}$}} & \cellcolor[gray]{0.9}{{$\boldsymbol{0.931_{(0.002)}}$}} & \cellcolor[gray]{0.9}{{$\boldsymbol{2.480_{(0.161)}}$}} & \cellcolor[gray]{0.9}{{$\boldsymbol{0.937_{(0.002)}}$}} & \cellcolor[gray]{0.9}{{$\boldsymbol{0.884_{(0.003)}}$}} & \cellcolor[gray]{0.9}{{$\boldsymbol{5.042_{(0.168)}}$}} & \cellcolor[gray]{0.9}{{$\boldsymbol{0.972_{(0.001)}}$}} & \cellcolor[gray]{0.9}{{$\boldsymbol{0.978_{(0.001)}}$}} & \cellcolor[gray]{0.9}{{$\boldsymbol{1.620_{(0.036)}}$}}  \\
    \bottomrule  
  \end{tabular}
  \end{adjustbox}
  \caption{Comparison with SOTAs on public datasets. $^{\dagger}:$ Note that SSL methods are tested under 50\% annotations. The best and second-best results are highlighted in \textbf{bold} and \underline{underlined}, respectively. $^{*}: p < 0.01$ comparing against our method in each metric. Moreover, we test the generalization capability of DeNAS-ViT on additional datasets of other fields, shown in Sec.~\ref{sec_generalization_capability}.}\label{comparison}
\end{table*}

The total loss function of our DeNAS-ViT is:
\begin{align}
\mathcal{L}_{total,a}=\lambda_{1}\mathcal{L}_{sup,a}+\lambda_{2}\mathcal{L}_{uns,a}+\lambda_{3}\mathcal{L}_{ind,a}+\lambda_{4}\mathcal{L}_{con,a},
\end{align}
where $\mathcal{L}_{sup,a}=\frac{1}{2}[CE(p_{l,a},y_{l,a})+DICE(p_{l,a},y_{l,a})]$ is the supervised loss, which is the combination of the cross-entropy loss and Dice loss calculated on the labeled dataset; $\mathcal{L}_{uns,a}$ refers to the unsupervised loss in Eq.~(\ref{formula_uns}) calculated on the unlabeled dataset; $\mathcal{L}_{ind,a}$ refers to the network independence loss in Eq.~(\ref{formula:ind}) based on the network architecture; $\mathcal{L}_{con,a}$ refers to the contrastive loss in Eq.~(\ref{formula:con}) based on the multi-resolution features. $\lambda_{1}$, $\lambda_{2}$, $\lambda_{3}$, and $\lambda_{4}$ are hyper-parameters to balance the relationship between losses.
\vspace{-1mm}
\subsection{Stage-wise Optimization Strategy}\label{optimization_strategy}
To adapt to the hybrid constraint-driven SSL, we propose a stage-wise optimization strategy, summarized in {Algorithm} \ref{alg1}. In each iteration, we first fix the network parameters and optimize the combination matrix for $E_{B}$ epochs. In the second stage, we fix the combination matrix and network architecture parameters and then update the network weights by minimizing $\mathcal{L}_{total}$. In the third stage, we only update the architecture parameters by minimizing $\mathcal{L}_{total}$ after $E_{A}$ epochs. During the optimization, continuous relaxation is implemented for the gradient descent algorithm.
\vspace{-1mm}
\begin{algorithm}[ht]
	\renewcommand{\algorithmicrequire}{\textbf{Input:}}
	\renewcommand{\algorithmicensure}{\textbf{Output:}}
    \caption{Optimization Strategy}
    \label{alg1}
    \begin{algorithmic}[1] 
    \REQUIRE Datasets $\mathcal{D}_{l}$, $\mathcal{D}_{u}$, weights $w$, architecture $\alpha,\beta,\gamma$, combination matrices $G_{a},G_{b}$, epochs $E$, $E_{A}$, and $E_{B}$.
    \ENSURE Searched $f(\theta_{a}^{*}), f(\theta_{b}^{*})$.
		\FOR{$e=1,\cdots,E$}{
            \FOR{$f=1,\cdots,E_{B}$}
            \STATE Fix $w,\alpha,\beta,\gamma$. Update $G_{a},G_{b}$ by maximizing $\mathcal{L}_{ind,a}$ and $\mathcal{L}_{ind,b}$, respectively.
            \ENDFOR
            \STATE Fix $G_{a},G_{b},\alpha,\gamma$. Update $w, \beta$ of $f(\theta_{a})$ and $f(\theta_{b})$ by minimizing $\mathcal{L}_{total,a}$ and $\mathcal{L}_{total,b}$, respectively. 
            \IF{$e>E_{A}$}{
            \STATE Fix $G_{a},G_{b},w,\beta$. Update $\alpha, \gamma$ of $f(\theta_{a})$ and $f(\theta_{b})$ by minimizing $\mathcal{L}_{total,a}$ and $\mathcal{L}_{total,b}$, respectively. }
            \ENDIF}
            \ENDFOR
    \end{algorithmic}
\end{algorithm}

\begin{table*}[ht]
  \centering
  \begin{threeparttable}
  \begin{adjustbox}{width=0.99\linewidth}
\begin{tabular}{cccccccc}
    \toprule
    Type & {Method} & Usage Variants & DSC$\uparrow$ & {IoU$\uparrow$} & {95HD$\downarrow$} & Params (M)$\downarrow$ & FLOPs (G)$\downarrow$ \\ 
    \midrule
    Manual & TransUNet  & Employ Transformer as Encoder & $0.906_{(0.003)}*$ & $0.905_{(0.004)}*$ & $4.032_{(0.207)}*$ & \phantom{0}96.07 & \textbf{\phantom{0}48.34} \\
    Manual & EfficientViT-L2  & Multi-scale Tokens & $0.916_{(0.005)}*$ & $0.915_{(0.004)}*$ & $3.356_{(0.260)}*$ & \phantom{0}52.12 & \phantom{0}91.45 \\
    Manual & ViT-H Med-SAM  & Employ Transformer as Encoder & \underline{$0.920_{(0.001)}*$} & \underline{$0.918_{(0.001)}*$} & \underline{$3.093_{(0.107)}*$} & 636.00 & 246.20 \\
    NAS & Auto-DeepLab  & No Transformer is Applied & $0.908_{(0.004)}*$ & $0.907_{(0.004)}*$ & $3.857_{(0.164)}*$ & \phantom{0}44.42 & 347.52 \\
    NAS & SSHNN$^{\dagger}$  & Treat Transformer as Additional Branch & $0.906_{(0.002)}*$ & $0.904_{(0.002)}*$ & $4.011_{(0.145)}*$ & \underline{\phantom{0}38.82} & \underline{\phantom{0}52.78} \\
    NAS & DAST  & Treat Transformer as Candidate Operation & $0.915_{(0.002)}*$ & $0.913_{(0.002)}*$ & $3.438_{(0.119)}*$ & 192.44 & 110.36 \\
    \midrule
    \rowcolor[gray]{0.95}
    NAS & \textbf{DeNAS-ViT-E}$^{\dagger}$ & Treat EfficientViT as Candidate Operation & $0.923_{(0.001)}*$ & $0.922_{(0.002)}*$ & $2.884_{(0.123)}$ & \textbf{\phantom{0}38.48} & \phantom{0}55.47 \\
   \rowcolor[gray]{0.9}
    NAS & \textbf{DeNAS-ViT}$^{\dagger}$ & Efficient NAS-ViT & {$\boldsymbol{0.933_{(0.002)}}$} & {$\boldsymbol{0.931_{(0.002)}}$} & {$\boldsymbol{2.480_{(0.161)}}$} & \phantom{0}41.20 & \phantom{0}67.50 \\
    \bottomrule
  \end{tabular}
     \end{adjustbox}
    \end{threeparttable}
    
    \caption{Discussion of various Transformer usage variants. DeNAS-ViT achieves the highest accuracy while maintaining no increase in parameter size (Params) or computation complexity (FLOPs). $^{\dagger}:$ SSL methods are tested under 50\% annotations.}\label{ablation_different}
  \end{table*}

\vspace{-2mm}
\section{Experiments}
\subsection{Datasets}
Three public datasets are utilized for evaluation: (i) {CAMUS dataset}~\cite{leclerc2019deep} is a large-scale 2D echocardiography dataset, comprising 2000 labeled images, and approximately 19000 unlabeled images. It includes four classes of labels: left ventricle endocardium (LV), left atrium, myocardium, and background. (ii) {HMC-QU dataset}~\cite{kiranyaz2020left} consists of 2D echocardiography videos. By splitting these sequences into individual images, a total of 4989 images are obtained, among which 2349 images are annotated with two classes of labels: left ventricle wall and background. (iii) {CETUS dataset}~\cite{bernard2015standardized} comprises 90 sequences of 3D ultrasound volumes. After randomly selecting 80 frames from each sequence, 7200 annotated images are obtained, with two classes of labels: LV and background. In our experiments, 3400 of these images are utilized as unlabeled data. More details on dataset preparation are provided in Appx.~\ref{dataset details}. 

\vspace{-1mm}

\subsection{Experimental Settings}
\noindent \textbf{Network Architecture.}
In the encoder NAS, each cell has $N=5$ intermediate nodes. For the resolution $r=4$, the channel numbers are fixed as 8 for each node. When $r$ doubles, the number of channels doubles accordingly.
For partial channel connections, $\frac{1}{4}$ of the channels are allowed for searching.
The default number of layers $L$ is set to 8 and the default proportion of labeled data utilized is 50\%.

\noindent \textbf{Training Setup.}
The hyper-parameters $\lambda_{1}=1$, $\lambda_{2}=\lambda_{4}=5exp(-5(1-\frac{min(i,I_{ramp})}{I_{ramp}})^{2})$ are adopted following~\cite{yu2019uncertainty, huang2023semi} at the $i$-th epoch, where $I_{ramp}=50$, and $\lambda_{3}=2$. The linear coefficient matrices $G$ are optimized by Adam with a fixed learning rate of 0.001. The network weights $w$ and architecture $\beta$ are optimized using SGD with an initial learning rate of 0.001, a momentum of 0.9, and a weight decay of 0.0003. For the architecture $\alpha$ and $\gamma$, the Adam optimizer is applied with a learning rate of 0.003 and weight decay of 0.001. The total number of epochs is set to $E=40$ and the architecture optimization begins at $E_{A}=10$. In each epoch, $G$ is updated $E_{B}=6$ times. Experiments are conducted on an Nvidia A100 GPU and are repeated with 5 random seeds, where mean value and standard deviation are reported in $mean_{(std)}$ format.

\noindent \textbf{Metrics.}
Dice Similarity Coefficient (DSC), Intersection over Union (IoU), and 95\% Hausdorff Distance (95HD). Wilcoxon signed-rank test for statistical significance.

\subsection{Main Results}\label{sec:comparison with SOTA}
\noindent \textbf{Comparison with SOTA Methods.}
We evaluate DeNAS-ViT against other SOTAs, including supervised handcrafts: UNet++, nnU-Net~\cite{isensee2021nnu}, Transfuse~\cite{zhang2021transfuse}; supervised NAS: Auto-DeepLab, M\textsuperscript{3}NAS~\cite{lu2022m}, GeNAS~\cite{jeong2023genas}; semi-supervised handcrafts: URPC~\cite{luo2021efficient}, CPS, CnT-B~\cite{huang2023semi}, ARCO-SG~\cite{you2024rethinking}; and semi-supervised NAS: Se\textsuperscript{2}NAS~\cite{pauletto2022se}, SSHNN~\cite{chen2023sshnn}, implemented by their officially released code and pre-trained model. 

{Table}~\ref{comparison} and Appx. Fig.~\ref{comparison_result} provide quantitative and qualitative comparisons across HMC-QU, CAMUS, and CETUS datasets. Supervised methods were trained on 100\% annotations, while semi-supervised learning (SSL) methods used 50\% annotated data, excluding unlabeled data. DeNAS-ViT consistently outperforms all methods across all metrics. On HMC-QU, it surpasses the previous SOTA (CnT-B) by 2.2\% in DSC, 2.3\% in IoU, and 1.622 in 95HD, while also demonstrating superior results on CAMUS and CETUS.

\noindent \textbf{Comparison with ViT Usage Variants.}
Table~\ref{ablation_different} provides an evaluation of SOTA models utilizing different Transformer application strategies on the HMC-QU dataset. TransUNet and Med-SAM~\cite{ma2024segment} utilize ViTs as their encoders. EfficientViT-L2~\cite{cai2023efficientvit} incorporates fixed multi-scale tokens within the ViT architecture. Auto-DeepLab, a NAS model, does not leverage Transformers. SSHNN integrates ViTs as an auxiliary branch, while DAST employs ViTs as a candidate operation at the feature scale. In our DeNAS-ViT-E, we also explored a variant where EfficientViT was used as a candidate operation instead of the Efficient NAS-ViT. Our approach achieves the highest accuracy with a smaller model size and reduced FLOPs. Furthermore, the DeNAS-ViT-E variant demonstrates that the choice of Transformer application significantly influences performance, providing deeper insights into our design.

\begin{figure}[ht]
    \centering
    \includegraphics[width=0.99\linewidth]{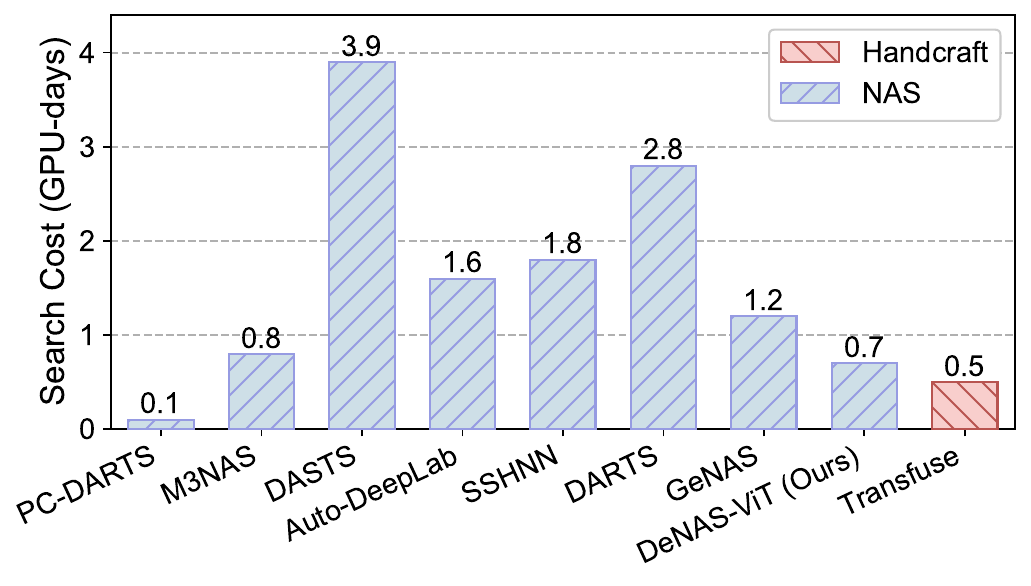}
    \vspace{-3mm}
    \caption{Search cost comparison with SOTA methods.}\label{cost_day}
\end{figure}
\noindent \textbf{Search Cost.}~Fig.~\ref{cost_day} compares the architecture search time of DeNAS-ViT with Transfuse (a high-performing handcrafted model, see Table~\ref{comparison}) and other NAS frameworks on the HMC-QU dataset. For Transfuse, the search cost reflects its training time, while for NAS models, it encompasses both search and training durations, measured on an Nvidia A100 GPU. DeNAS-ViT requires 0.7 GPU days, with its search process being notably efficient and competitive compared to handcrafted models like Transfuse.

\subsection{Robustness and Generalization}\label{sec_generalization_capability}
\begin{figure}[t]
    \centering
    \includegraphics[width=0.99\linewidth]{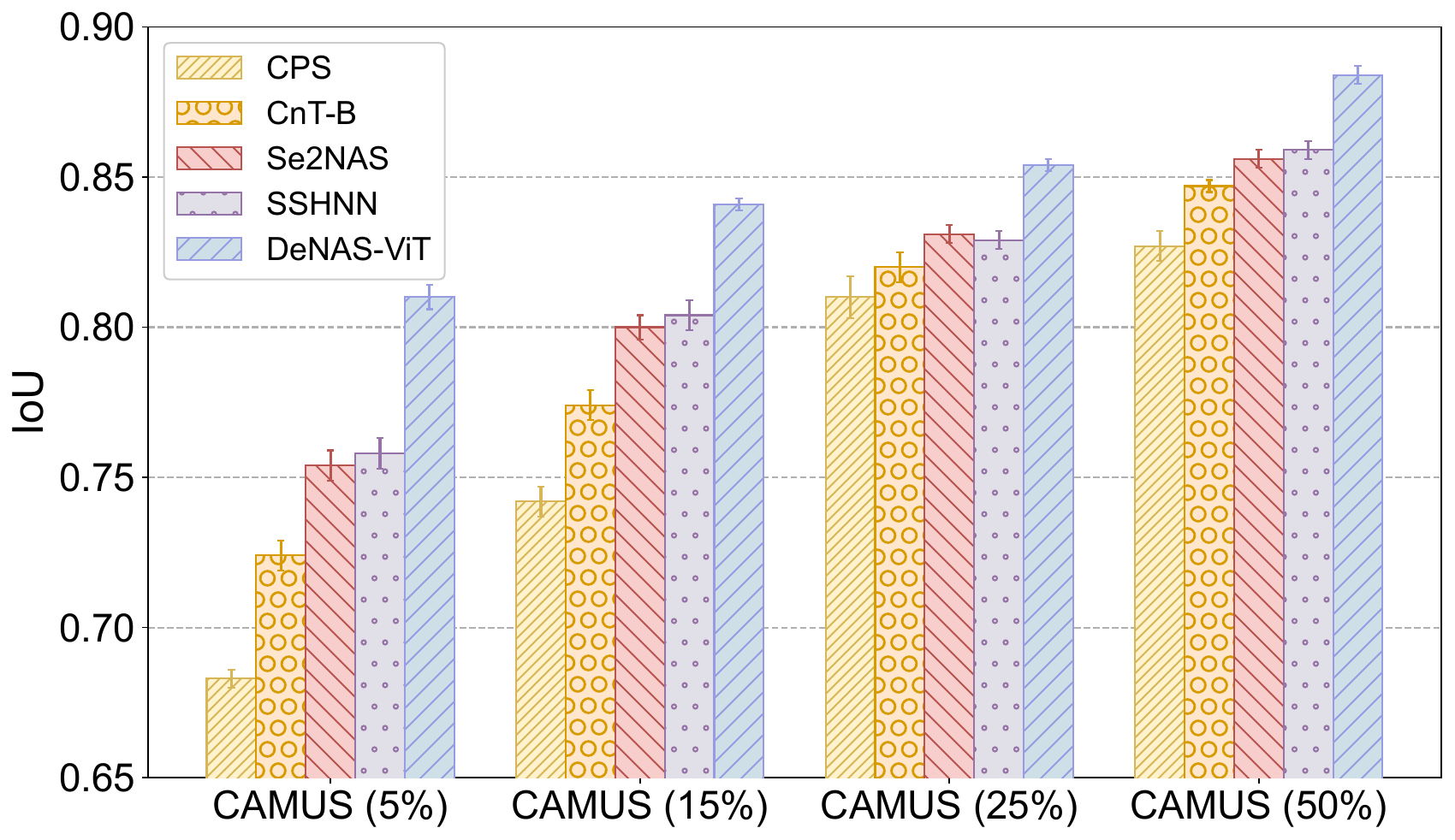}
    \caption{The impact of annotation proportions on SOTAs is evaluated using the CAMUS dataset, where DeNAS-ViT exhibits both robustness and effectiveness.}\label{loss_compare}
\end{figure}

\noindent \textbf{Robustness on Ultrasound Images.}
Fig.~\ref{loss_compare} illustrates the results of representative SSL networks and DeNAS-ViT with varying proportions of annotations on the CAMUS dataset. It can be observed that DeNAS-ViT performs more stably than other advanced SSL methods. When the annotation ratio descends from 50\% to 5\%, the IoU of CPS, CnT-B, Se\textsuperscript{2}NAS, and SSHNN drops by 14.4\%, 12.3\%, 10.2\%, and 10.1\%, respectively. In contrast, DeNAS-ViT’s IoU drops by only 7.4\%, highlighting its robustness. This stability is attributed to its hybrid constraint-driven SSL, which improves data utilization.

\begin{table}[ht]
  \centering
  \begin{adjustbox}{width=\linewidth}
  \begin{threeparttable}
    \begin{tabular}{c|cccc}
    \toprule
    \multirow{2}{*}{Method} & \multicolumn{2}{c}{ISIC (10\%)} & \multicolumn{2}{c}{ISIC (15\%)} \\ 
     & DSC$\uparrow$ & 95HD$\downarrow$ & DSC$\uparrow$ & 95HD$\downarrow$ \\
    \midrule
    CPS & $0.746$ & $15.722$ & $0.779$ & $12.910$  \\
    CnT-B  & $0.790$ ({\textcolor{iccvblue}{+0.044}}) & $11.879$ ({\textcolor{iccvblue}{-3.843}}) & ${0.814}$ ({\textcolor{iccvblue}{+0.035}})  & \underline{$9.583$} ({\textcolor{iccvblue}{-3.327}}) \\
    ARCO-SG  & \underline{$0.798$} ({\textcolor{iccvblue}{+0.052}}) & \underline{$11.352$} ({\textcolor{iccvblue}{-4.370}}) & \underline{$0.816$} ({\textcolor{iccvblue}{+0.037}}) & $9.600$ ({\textcolor{iccvblue}{-3.310}}) \\
    SSHNN & $0.788$ ({\textcolor{iccvblue}{+0.042}})& $12.064$ ({\textcolor{iccvblue}{-3.658}})& ${0.809}$ ({\textcolor{iccvblue}{+0.030}})& $10.119$ ({\textcolor{iccvblue}{-2.791}}) \\
    \rowcolor[gray]{0.9}
    \textbf{DeNAS-ViT}  & $\boldsymbol{0.817}$ (\textbf{\textcolor{iccvblue}{+0.071}})  & $\boldsymbol{9.024}$ (\textbf{\textcolor{iccvblue}{-6.698}}) & $\boldsymbol{0.840}$ (\textbf{\textcolor{iccvblue}{+0.061}}) & $\boldsymbol{7.391}$ (\textbf{\textcolor{iccvblue}{-5.519}}) \\
    \bottomrule
    \end{tabular} 
    \end{threeparttable}
    \end{adjustbox}
    \caption{Comparison with SOTA methods on the ISIC under varying annotation ratios to assess the generalizability.}\label{main:generalization}
  \end{table}

\noindent \textbf{Generalization Across Image Domains.} To test the generalization capability of DeNAS-ViT, we conduct experiments on the International Skin Imaging Collaboration (ISIC)~\cite{codella2018skin} dataset.
As shown in Table \ref{main:generalization}, we evaluate our method against other SOTAs, including CPS, CnT-B, ARCO-SG, and SSHNN. Our method achieves the best performance in all semi-supervised settings (10\% and 15\%), with the improvements of DSC: 1.9\%, 2.4\%, and 95HD: -2.328mm, -2.209mm over the runner-up. Extensive results demonstrate the generalization capability of our proposed model. More experiments are shown in Sec.~\ref{sec:generalization_tests}.

\subsection{Ablation Studies}\label{ablation}
\noindent \textbf{Effects of Sub-modules.}
To evaluate the impact of sub-modules, ablation studies are conducted as shown in Tab.~\ref{ablation_1}. The baseline model (No. 1) is the NAS backbone (Sec.~\ref{NAS backbone}) without the Efficient NAS-ViT module, trained on 50\% labeled data. Adding the Efficient NAS-ViT module (No. 2) improves segmentation performance, increasing the DSC from 0.905 to 0.912, demonstrating enhanced context extraction capability. Introducing the co-training SSL framework (No. 6) further boosts performance by 0.7\%, attributed to the additional information gained from unlabeled data. Incorporating network independence loss (No. 7) results in a 9.5\% improvement in 95HD over No. 6. The effectiveness of contrastive learning is validated in No. 5, which achieves a 0.7\% higher DSC compared to using alone $\mathcal{L}_{uns}$ in No. 3.

\begin{table}[ht]
  \centering
  \begin{threeparttable}
  \begin{adjustbox}{width=0.99\linewidth}
  \begin{tabular}{ccccc|ccc}
    \toprule
    {No.} & {NAS} & {$\mathcal{L}_{uns}$} & {$\mathcal{L}_{ind}$} & {$\mathcal{L}_{con}$} & DSC$\uparrow$ & IoU$\uparrow$ & 95HD$\downarrow$  \\ 
    \midrule
    1 & \XSolidBrush & \XSolidBrush & \XSolidBrush & \XSolidBrush & $0.905_{(0.002)}$ & $0.906_{(0.003)}$ & $4.310_{(0.204)}$    \\ 
    2 & \Checkmark & \XSolidBrush & \XSolidBrush & \XSolidBrush & $0.912_{(0.001)}$ & $0.911_{(0.002)}$ & $3.623_{(0.131)}$  \\ 
    3 & \XSolidBrush & \Checkmark & \XSolidBrush & \XSolidBrush & $0.910_{(0.001)}$ & $0.910_{(0.001)}$ & $3.674_{(0.100)}$   \\ 
    4 & \XSolidBrush & \Checkmark & \Checkmark & \XSolidBrush & $0.919_{(0.001)}$ & $0.920_{(0.002)}$ & $3.058_{(0.093)}$  \\ 
    5 & \XSolidBrush & \Checkmark & \XSolidBrush & \Checkmark & $0.917_{(0.003)}$ & $0.915_{(0.002)}$ & $3.356_{(0.297)}$ 
     \\ 
    6 & \Checkmark & \Checkmark & \XSolidBrush & \XSolidBrush & $0.919_{(0.002)}$ & $0.918_{(0.002)}$ & $3.041_{(0.165)}$ 
     \\ 
    7 & \Checkmark & \Checkmark & \Checkmark & \XSolidBrush & $0.926_{(0.002)}$ & $0.925_{(0.001)}$ & $2.752_{(0.102)}$ 
      \\
    \rowcolor[gray]{0.9}
    8 & \Checkmark & \Checkmark & \Checkmark & \Checkmark &{{$\boldsymbol{0.933_{(0.002)}}$}} & {{$\boldsymbol{0.931_{(0.002)}}$}} & {{$\boldsymbol{2.480_{(0.161)}}$}}    \\
    \bottomrule  
\end{tabular}
    \end{adjustbox}
    \end{threeparttable}
    \caption{Ablation studies of each component in the DeNAS-ViT structure on the HMC-QU dataset. ``NAS'' denotes our proposed Efficient NAS-ViT. More experiments on the other two datasets are shown in Appx.~\ref{app_sec:effects of submodules}.}\label{ablation_1}
  \end{table}

\begin{figure}[t]
    \centering
    \includegraphics[width=0.99\linewidth]{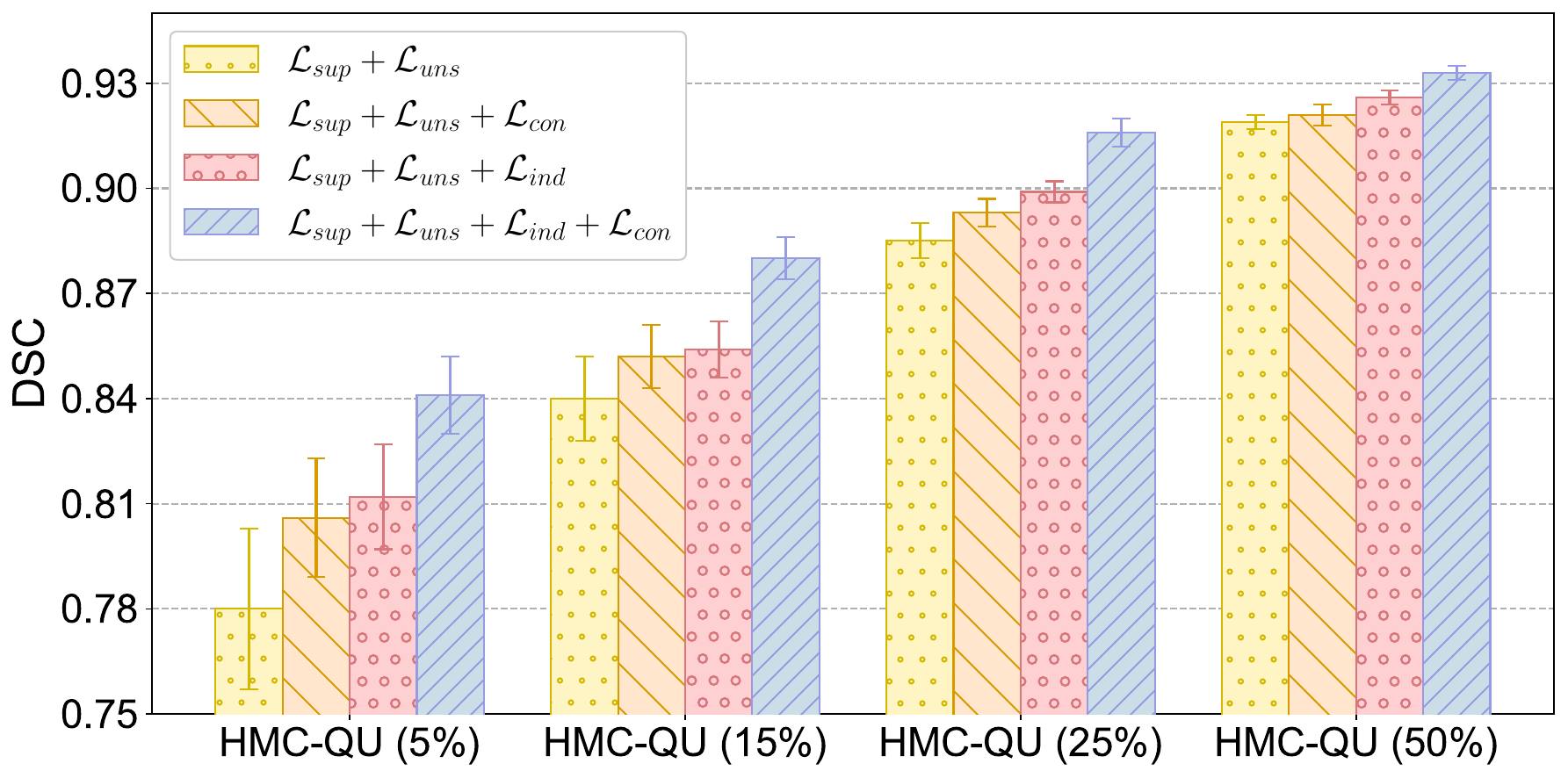}
    \vspace{-1mm}
    \caption{Sensitivity analysis of hybrid constraints. The case considering additional constraints shows better robustness.}\label{annotation}
\end{figure}

\noindent \textbf{Analysis of Hybrid Constraints.}
We investigate whether algorithmic independence loss $\mathcal{L}_{ind}$ or contrastive loss $\mathcal{L}_{con}$ would impact its performance under varying annotations on the HMC-QU, presented in Fig.~\ref{annotation}. When both $\mathcal{L}_{ind}$ and $\mathcal{L}_{con}$ are not considered, there is a 13.9\% drop in DSC as the annotation ratio decreases from 50\% to 5\%, which is larger than the 11.4\% and 11.5\% DSC decreases observed when $\mathcal{L}_{con}$ or $\mathcal{L}_{ind}$ is excluded, respectively. Notably, the DSC of DeNAS-ViT decreases by 9.2\%, representing the smallest drop among these cases. Results show that (1) $\mathcal{L}_{ind}$ encourages the creation of complementary networks that assist in training by providing multi-view information from the same data;
(2) $\mathcal{L}_{con}$ considers feature-level uncertainty where regions with lower uncertainty are filtered out, which is helpful for stable training under limited annotations. 

\vspace{-1mm}

\section{Conclusion}
In this paper, we propose a data efficient NAS-optimized ViT (DeNAS-ViT) to address two key challenges in ultrasound segmentation: multi-scale feature extraction and model robustness to limited data. For the first issue, DeNAS-ViT implements a three-level search, coupled with the Efficient NAS-ViT to enhance context extraction. For the second issue, we propose a NAS-based constraint-driven SSL for limited annotations. While experiments show DeNAS-ViT's effectiveness, our future direction is to enhance its zero-shot generalization capability across modalities.

\bibliography{aaai2026}

\newcommand{\answerYes}[1][]{\textcolor{blue}{[Yes] #1}}
\newcommand{\answerNo}[1][]{\textcolor{orange}{[No] #1}}
\newcommand{\answerNA}[1][]{\textcolor{gray}{[N/A] #1}}

\clearpage
\newpage
\renewcommand{\thesection}{\Alph{section}}
\renewcommand\thefigure{\Alph{section}\arabic{figure}} 
\renewcommand\thetable{\Alph{section}\arabic{table}}  
\setcounter{section}{0}
\setcounter{figure}{0} 
\setcounter{table}{0} 

\appendix

In the supplementary material, we provide additional details and experimental results to enhance the understanding and insights into our proposed DeNAS-ViT. This supplementary material is organized as:
\begin{itemize}[leftmargin=*]
\item In Sec.~\ref{more details}, a more detailed description of the role of architecture parameters for the NAS backbone. The candidate operations used in the encoder and decoder cell searches are shown.
\item In Sec.~\ref{sec:generalization_tests}, we test the generalization capability of our proposed HCS-TNAS compared with previous SOTA methods on two additional medical image datasets from another field.
\item In Sec.~\ref{app_sec:more experiments}, We provide more details about the employed datasets (Sec.~\ref{dataset details}) and a comprehensive analysis of the experimental results from the perspectives of visualization results (Sec.~\ref{app_sec:visualization}), training convergence (Sec.~\ref{sec:convergence}), and proportions of candidate operations in the final searched architecture (Sec.~\ref{analysis on candidate}), respectively, having a deeper insight into the structure design.
\item In Sec.~\ref{app_sec:more ablation studies}, we present additional ablation studies, analyzing the effects of each sub-module in DeNAS-ViT on more datasets (Sec.~\ref{app_sec:effects of submodules}), various candidate operation sets in the search process (Sec.~\ref{effect of candidate operation}), and different feature resolutions used in the training constraint (Sec.~\ref{impact of contrastive}), respectively.
\item In Sec.~\ref{hyper-parameter study}, we present additional hyper-parameter studies, discussing the impact of the loss coefficients (Sec.~\ref{app_sec:sensitivity to loss}) and the network size (Sec.~\ref{app_sec:sensitivity to layers}).
\item In Sec.~\ref{app_sec:limitations}, we discuss the potential for model improvements and our future research directions. 
\end{itemize}
 
\section{More Details about DeNAS-ViT}\label{more details}

\begin{table}[ht]
  \centering
  \begin{tabular}{ccc}
  \toprule
  Candidate Operations & Encoder & Decoder  \\ 
    \hline 
     {3$\times$3 separable convolution}  & {\Checkmark} & {\Checkmark}  \\    
     {5$\times$5 separable convolution}  & {\Checkmark} & {\Checkmark}  \\
   {3$\times$3 convolution with dilation rate 2}   & {\Checkmark} & {\Checkmark}  \\ 
   {5$\times$5 convolution with dilation rate 2}   & {\Checkmark} & {\Checkmark}  \\ 
   {3$\times$3 average/max pooling}   & {\Checkmark} & \XSolidBrush \\ 
   {skip connection}   & {\Checkmark} & \Checkmark  \\ 
   {no connection (zero)} & {\Checkmark} & \XSolidBrush \\
   \bottomrule
    \end{tabular} 
    \caption{Encoder cell and decoder cell candidate operations, where different candidate operations are considered according to their respective roles. The rationality of the design will be discussed in Sec. \ref{analysis on candidate}.}\label{Candidate Operations}
  \end{table}

As we introduced in the Sec. ``Methodology'' of the main paper, the architecture parameters $\alpha$ and $\gamma$ are responsible for updating the candidate operations in the encoder and decoder cells, respectively. In DeNAS-ViT, to enlarge the search space, unlike existing works where all cells share the same architecture parameters, our architecture parameters $\alpha$ for encoder cells are defined as a 3-dimensional tuple ($Cell, Edge, Operation$), and $\gamma$ for decoder cells is defined as a 2-dimensional tuple ($Cell, Operation$).
\begin{itemize}[leftmargin=*]
\item $Cell$ indicates that different cells are optimized by specialized parameters, overcoming the limitation of a simple cell type, and enabling a larger search space to be explored.
\item $Edge$ represents the connection between each intermediate node within a cell, which is used for feature aggregation at the intermediate nodes. Note that the decoder cell does not have this parameter to reduce the parameter count and training time. 
\item $Operation$ denotes the number of operation types. As shown in Table \ref{Candidate Operations}, 8 different choices are provided for the encoder to ensure diversity in the selection, while 5 different choices are provided for the decoder. The encoder and decoder have different sets of candidate operations. The rationale behind designing different candidate operation sets will be discussed in more ablation studies.
\end{itemize}
Thus, in the search process for the encoder and decoder cells, we allow different cells to have different structures, and different node-to-node connections to have different operations. This approach enables the network
to have a vast search space and optimization potential without requiring a large number of layers.

\begin{figure*}[t]
    \centering
    \includegraphics[width=\linewidth]{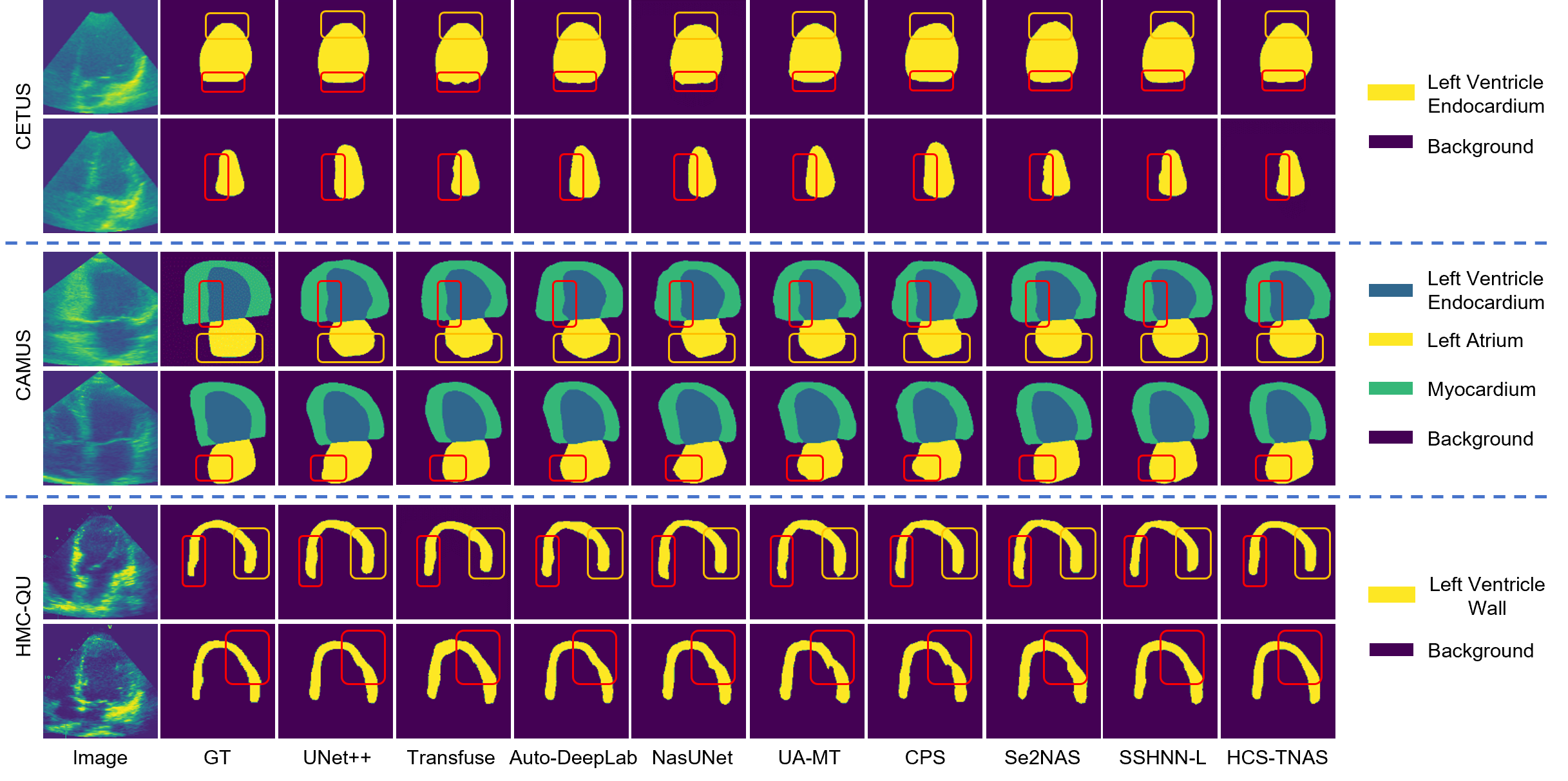}
    \caption{Visualization of segmentation results obtained by our proposed DeNAS-ViT and other SOTA methods on the CETUS, the CAMUS, and the HMC-QU datasets. Note that Resize and crop operations are applied to the model's predictions to visualize their segmentation differences. Red or orange boxes are used to highlight obvious wrong predictions, showing that our method has fewer over-segmentation and under-segmentation issues, yielding segmented shapes that closely match expert annotations.}\label{comparison_result}
\end{figure*}  

\begin{table}[ht]
  \centering
  \begin{adjustbox}{width=\linewidth}
  \begin{threeparttable}
    \begin{tabular}{c|cccc}
    \toprule
    \multirow{2}{*}{Method} & \multicolumn{2}{c}{ISIC (10\%)} & \multicolumn{2}{c}{ISIC (15\%)} \\ 
     & DSC$\uparrow$ & 95HD$\downarrow$ & DSC$\uparrow$ & 95HD$\downarrow$ \\
    \midrule
    CPS & $0.746$ & $15.722$ & $0.779$ & $12.910$  \\
    CnT-B  & $0.790$ ({\textcolor{iccvblue}{+0.044}}) & $11.879$ ({\textcolor{iccvblue}{-3.843}}) & ${0.814}$ ({\textcolor{iccvblue}{+0.035}})  & \underline{$9.583$} ({\textcolor{iccvblue}{-3.327}}) \\
    ARCO-SG  & \underline{$0.798$} ({\textcolor{iccvblue}{+0.052}}) & \underline{$11.352$} ({\textcolor{iccvblue}{-4.370}}) & \underline{$0.816$} ({\textcolor{iccvblue}{+0.037}}) & $9.600$ ({\textcolor{iccvblue}{-3.310}}) \\
    SSHNN & $0.788$ ({\textcolor{iccvblue}{+0.042}})& $12.064$ ({\textcolor{iccvblue}{-3.658}})& ${0.809}$ ({\textcolor{iccvblue}{+0.030}})& $10.119$ ({\textcolor{iccvblue}{-2.791}}) \\
    \rowcolor[gray]{0.9}
    \textbf{DeNAS-ViT}  & $\boldsymbol{0.817}$ (\textbf{\textcolor{iccvblue}{+0.071}})  & $\boldsymbol{9.024}$ (\textbf{\textcolor{iccvblue}{-6.698}}) & $\boldsymbol{0.840}$ (\textbf{\textcolor{iccvblue}{+0.061}}) & $\boldsymbol{7.391}$ (\textbf{\textcolor{iccvblue}{-5.519}}) \\
    \bottomrule
    \end{tabular} 
    \end{threeparttable}
    \end{adjustbox}
    \caption{Comparison with SOTA methods on the ISIC under varying annotation ratios to assess the generalizability.}\label{main:generalization_1}
  \end{table}

\begin{table}[ht]
  \centering
  \begin{adjustbox}{width=\linewidth}
  \begin{threeparttable}
    \begin{tabular}{c|cccc}
    \toprule
    \multirow{2}{*}{Method}  & \multicolumn{2}{c}{ACDC (10\%)} & \multicolumn{2}{c}{ACDC (15\%)} \\ 
     & DSC$\uparrow$& 95HD$\downarrow$ & DSC$\uparrow$ & 95HD$\downarrow$\\
    \midrule
    CPS  & $0.831$ & $10.621$ & $0.851$ & $7.304$  \\
    CnT-B  & \underline{$0.876$} ({\textcolor{iccvblue}{+0.045}}) & \underline{$5.120$} ({\textcolor{iccvblue}{-5.501}}) & ${0.894}$ ({\textcolor{iccvblue}{+0.043}}) & {$2.853$} ({\textcolor{iccvblue}{-4.451}}) \\
    ARCO-SG  & {$0.872$} ({\textcolor{iccvblue}{+0.041}}) & {$6.144$} ({\textcolor{iccvblue}{-4.477}}) & \underline{$0.896$} ({\textcolor{iccvblue}{+0.045}}) & \underline{$2.607$} ({\textcolor{iccvblue}{-4.697}}) \\
    SSHNN  & $0.868$ ({\textcolor{iccvblue}{+0.037}})& $6.722$ ({\textcolor{iccvblue}{-3.899}})& ${0.889}$ ({\textcolor{iccvblue}{+0.038}}) & $3.519$ ({\textcolor{iccvblue}{-3.785}}) \\
    \rowcolor[gray]{0.9}
    \textbf{DeNAS-ViT}  & $\boldsymbol{0.890}$ ({\textbf{\textcolor{iccvblue}{+0.059}}})& $\boldsymbol{3.356}$ (\textbf{\textcolor{iccvblue}{-7.265}})& $\boldsymbol{0.915}$ (\textbf{\textcolor{iccvblue}{+0.064}}) & $\boldsymbol{1.677}$ (\textbf{\textcolor{iccvblue}{-5.627}})\\
    \bottomrule
    \end{tabular} 
    \end{threeparttable}
    \end{adjustbox}
    \caption{Comparison with SOTA methods on the ACDC under varying annotation ratios to verify the generalizability.}\label{generalization_2}
  \end{table}

\section{Generalization Across Image Domains}\label{sec:generalization_tests} 
To test the generalization capability of DeNAS-ViT, we further conduct experiments on the International Skin Imaging Collaboration (ISIC)~\cite{codella2018skin} dataset and Automated Cardiac Diagnosis Challenge (ACDC)~\cite{bernard2018deep}. (1) The ISIC dataset is a skin lesion segmentation dataset comprising 2594 dermoscopy images, with 1838 training images and 756 validation images. (2) The ACDC dataset is a magnetic resonance imaging scan dataset that includes 100 scans of three organs. Following \cite{huang2023semi}, we use 70 samples for training and 30 samples for testing. All images are resized to 224$\times$224 pixels. For a semi-supervised setting, 10\% and 15\% training data are labeled, while the rest training data are unlabeled.

As shown in Table \ref{main:generalization_1} and Table \ref{generalization_2}, we evaluate our method against other SOTAs, including CPS~\cite{chen2021semi}, CnT-B~\cite{huang2023semi}, ARCO-SG~\cite{you2024rethinking}, and SSHNN~\cite{chen2023sshnn}. Our method achieves the best segmentation performance in all semi-supervised settings (10\% and 15\%), on the ICIC dataset, with the improvements of DSC: 1.9\%, 2.4\%, and 95HD: -2.328mm, -2.209mm over the runner-up; on the ACDC dataset, with the improvements of DSC: 1.4\%, 1.9\%, and 95HD: -1.764mm, -0.930mm over the runner-up. Extensive results demonstrate the generalization capability of our proposed model.

\section{More Experimental Results}\label{app_sec:more experiments}
\subsection{Dataset Details}\label{dataset details}
\subsubsection{CAMUS Dataset.} The CAMUS dataset~\cite{leclerc2019deep} is a large open 2D echocardiography dataset collected from 500 patients. Each patient contributes a 4-chamber and a 2-chamber view sequence, each containing approximately 20 unlabeled images. Exceptions are made for annotations at the moments of end-diastole (ED) and end-systole (ES). Thus, there are 2000 labeled images and approximately 19000 unlabeled images. The segmentation labels include four categories: the left ventricle endocardium, the myocardium, the left atrium, and the background. All images are resized to 224$\times$224 pixels for training.

\subsubsection{HMC-QU Dataset.} The HMC-QU dataset~\cite{kiranyaz2020left} is composed of 2D echocardiography videos from the apical 4-chamber (A4C) and apical 2-chamber (A2C) views. In a specific subset of the HMC-QU dataset, 109 A4C view echocardiography recordings from 72 myocardial infarction (MI) patients and 37 non-MI subjects are annotated for one cardiac cycle. The remaining frames are unlabeled. Through the statistical analysis, the HMC-QU dataset consists of 4989 images in total, with 2349 images labeled for two categories: the left ventricle wall and the background. All images are resized to 192$\times$192 pixels for training.

\subsubsection{CETUS Dataset.} The CETUS dataset~\cite{bernard2015standardized} comprises 90 sequences (both ED and ES) of 3D ultrasound volumes capturing one complete cardiac cycle, collected from 45 patients. It can be categorized into three subgroups: 15 healthy individuals, 15 patients with a history of myocardial infarction at least 3 months prior to data acquisition, and 15 patients with dilated cardiomyopathy. To address the non-uniform duration of left ventricle endocardium presence and ensure fairness, we randomly selected 80 frames from each sequence to form the dataset. Consequently, the selection process provides 7200 labeled images, all resized to 192$\times$192 pixels. The ground-truth mask consists of two distinct labels: the left ventricle endocardium and the background. 

\subsubsection{Dataset Partitioning Approach.} For the CAMUS dataset, the training and test sets comprise 1800 and 200 labeled images, respectively, while for the CETUS dataset, they comprise 3400 and 400 labeled images, and for the HMC-QU dataset, they comprise 2000 and 349 labeled images. For the SSL experiments with DeNAS-ViT on the CAMUS dataset, we randomly selected 90, 270, 450, and 900 images from the training set, corresponding to 5\%, 15\%, 25\%, and 50\% labeled data, respectively, while using 19432 unlabeled images for training. On the CETUS dataset, we randomly selected 170, 510, 850, and 1700 images from the training set, corresponding to 5\%, 15\%, 25\%, and 50\% labeled data, respectively, and used 3400 unlabeled images for training. For the HMC-QU dataset, 100, 300, 500, and 1000 images are randomly selected as 5\%, 15\%, 25\%, and 50\% labeled data, respectively, while 2640 unlabeled images are used for training.

\subsection{Visualization Results}\label{app_sec:visualization}
{Fig.} \ref{comparison_result} elaborates the qualitative results of our proposed DeNAS-ViT against previous SOTA methods on the CETUS, the CAMUS, and the HMC-QU datasets. It can be observed that our method mitigates over and under-segmentation issues, and exhibits smoother boundaries between different labels. Overall, our proposed DeNAS-ViT achieves better segmentation results.

\begin{figure}[t]
    \centering
    \includegraphics[width=0.99\linewidth]{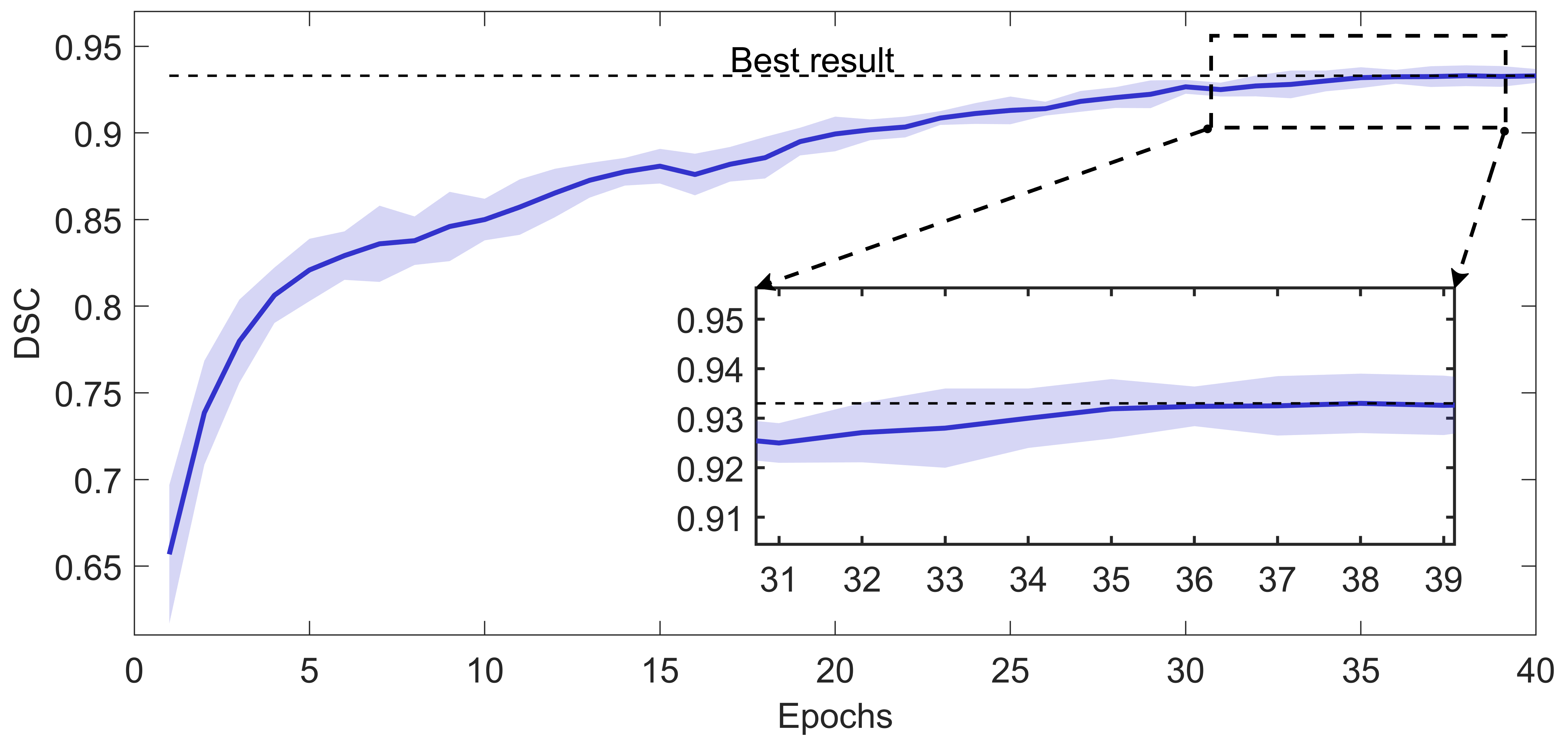}
    \caption{Validation accuracy during 40 epochs of architecture search optimization across 5 random trials. It can be seen that the curve steadily rises and can converge.}\label{dsc_curve}
\end{figure}

\begin{figure}[t]
    \centering
    \includegraphics[width=0.99\linewidth]{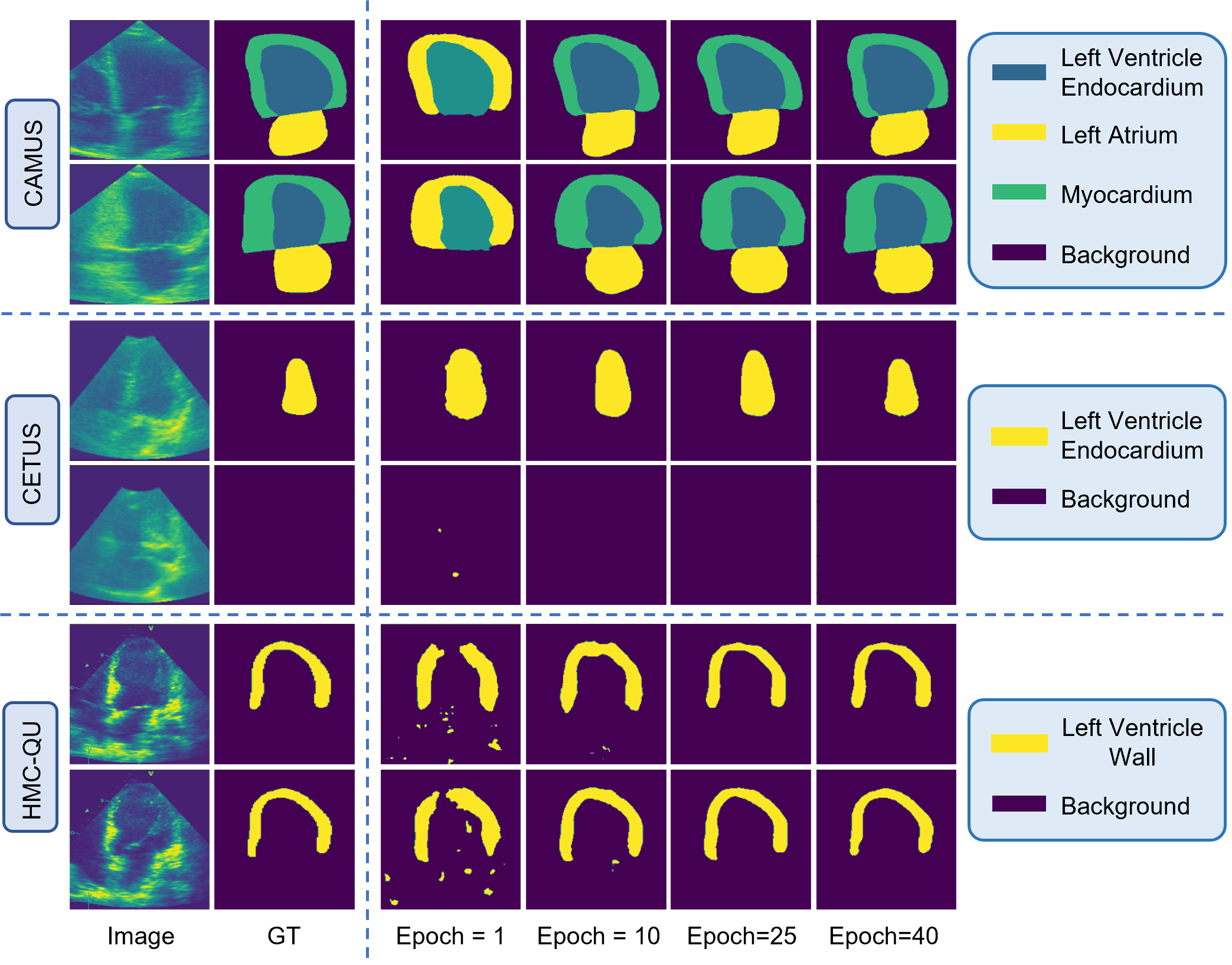}
    \caption{Qualitative assessment of the DeNAS-ViT throughout training process. As training iterations increases, results gradually approach expert annotations.}\label{epochs_result}
\end{figure}

\subsection{Convergence Behavior}\label{sec:convergence}
Fig.~\ref{dsc_curve} illustrates the trajectory of validation accuracy, measured in terms of DSC, across 40 epochs during the architecture search optimization process.
We report experimental results from 5 random trials on the HMC-QU dataset using 50\% labeled data. The deep blue curve represents the mean validation accuracy, while the light blue shaded region indicates variability across trials. The validation accuracy shows an initial steep rise, followed by a gradual increase, eventually converging near 0.933. An inset highlights the curve's behavior in the final epochs, revealing minor fluctuations within a narrow range and demonstrating stability after 37 epochs. Extending the search to longer epochs (50, 60, 70) did not yield significant improvements.
Fig.~\ref{epochs_result} depicts the progressive refinement of segmentation predictions during the training of DeNAS-ViT on the CAMUS, CETUS, and HMC-QU datasets, illustrating the model's continuous improvement in this clinically relevant task.

\subsection{Analysis on Candidate Operation Proportion}\label{analysis on candidate}
\begin{figure}[H]
    \centering
    \includegraphics[width=0.86\linewidth]{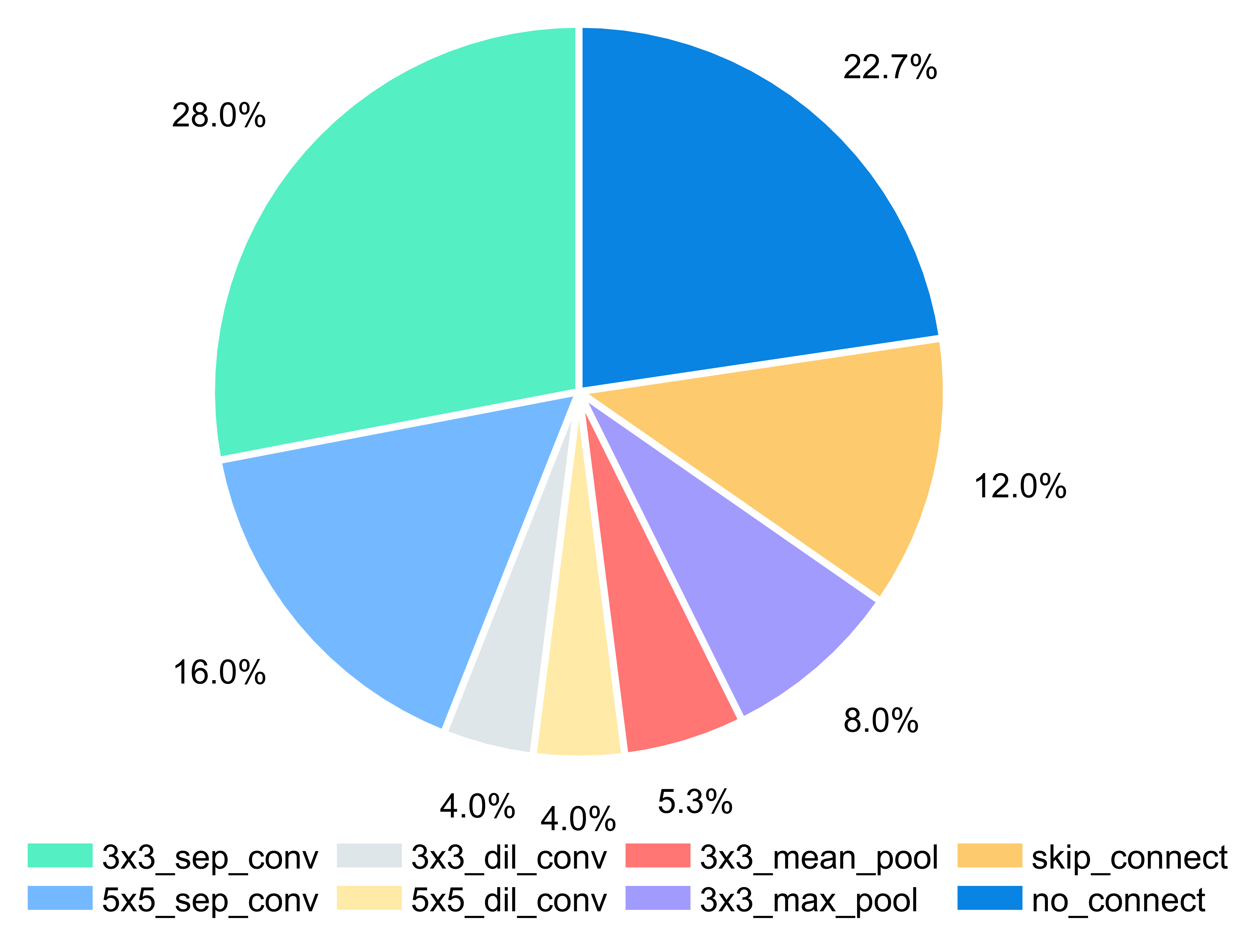}
    \caption{The proportion of each candidate operation in the cells of the encoder NAS of the final searched network on the HMC-QU dataset.}\label{encoder_2}
\end{figure}

\begin{figure}[H]
    \centering
    \includegraphics[width=0.86\linewidth]{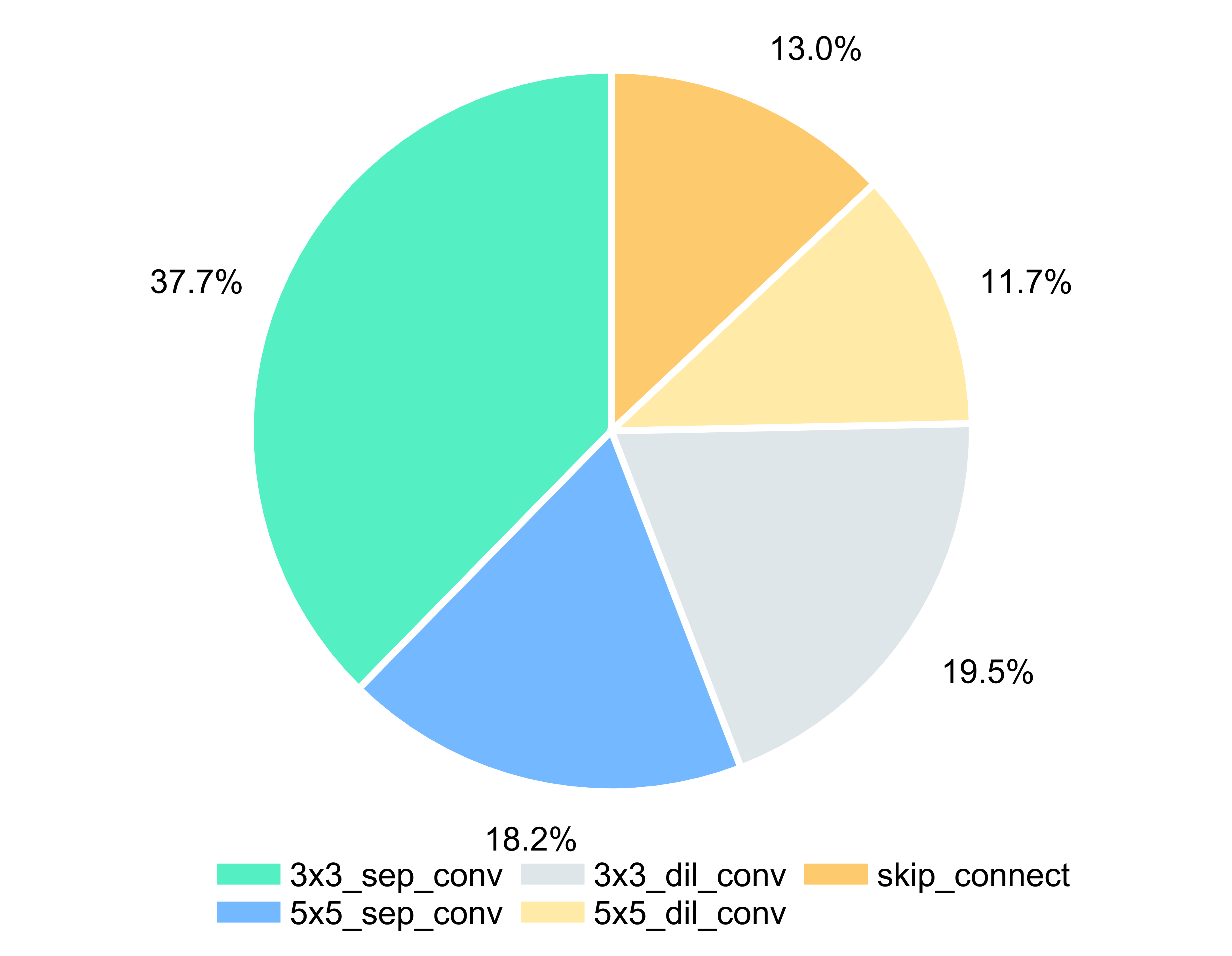}
    \caption{The proportion of each candidate operation in the cells of the decoder NAS of the final searched network on the HMC-QU dataset.}\label{decoder_2}
\end{figure}

\begin{figure}[H]
    \centering
    \includegraphics[width=0.86\linewidth]{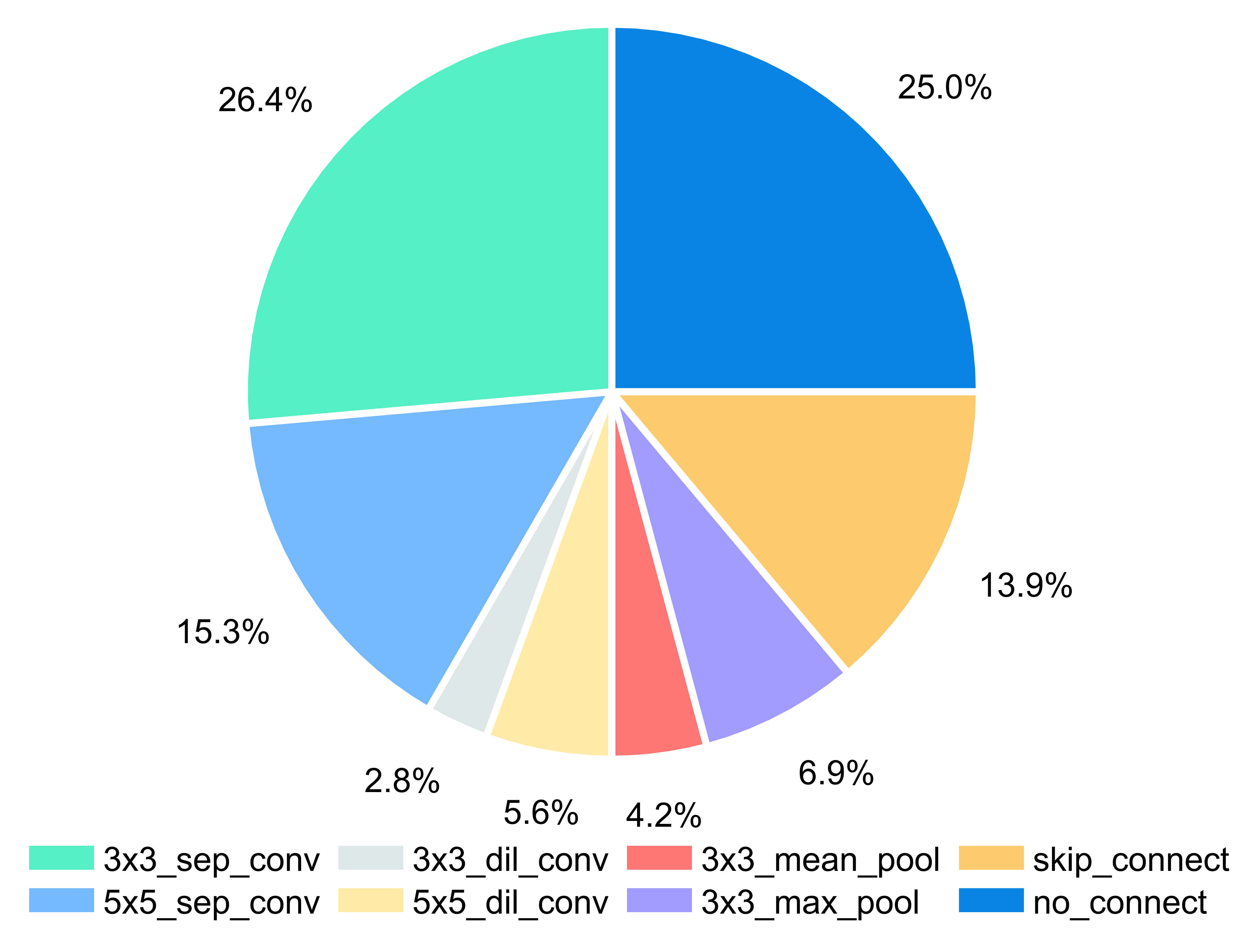}
    \caption{The proportion of each candidate operation in the cells of the encoder NAS of the final searched network on the CAMUS dataset.}\label{encoder}
\end{figure}

\begin{figure}[H]
    \centering
    \includegraphics[width=0.86\linewidth]{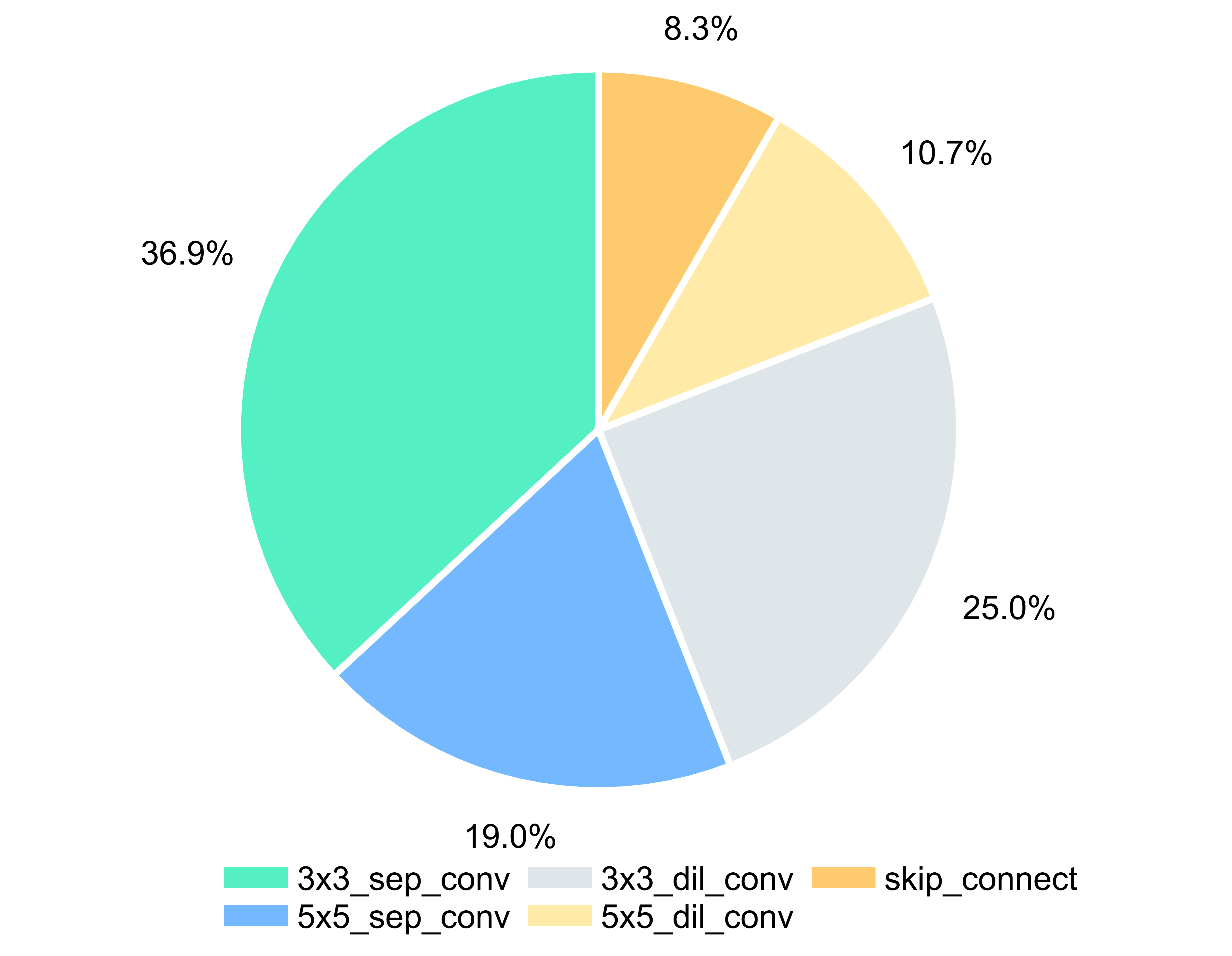}
    \caption{The proportion of each candidate operation in the cells of the decoder NAS of the final searched network on the CAMUS dataset.}\label{decoder}
\end{figure}

\begin{figure}[ht]
    \centering
    \includegraphics[width=0.86\linewidth]{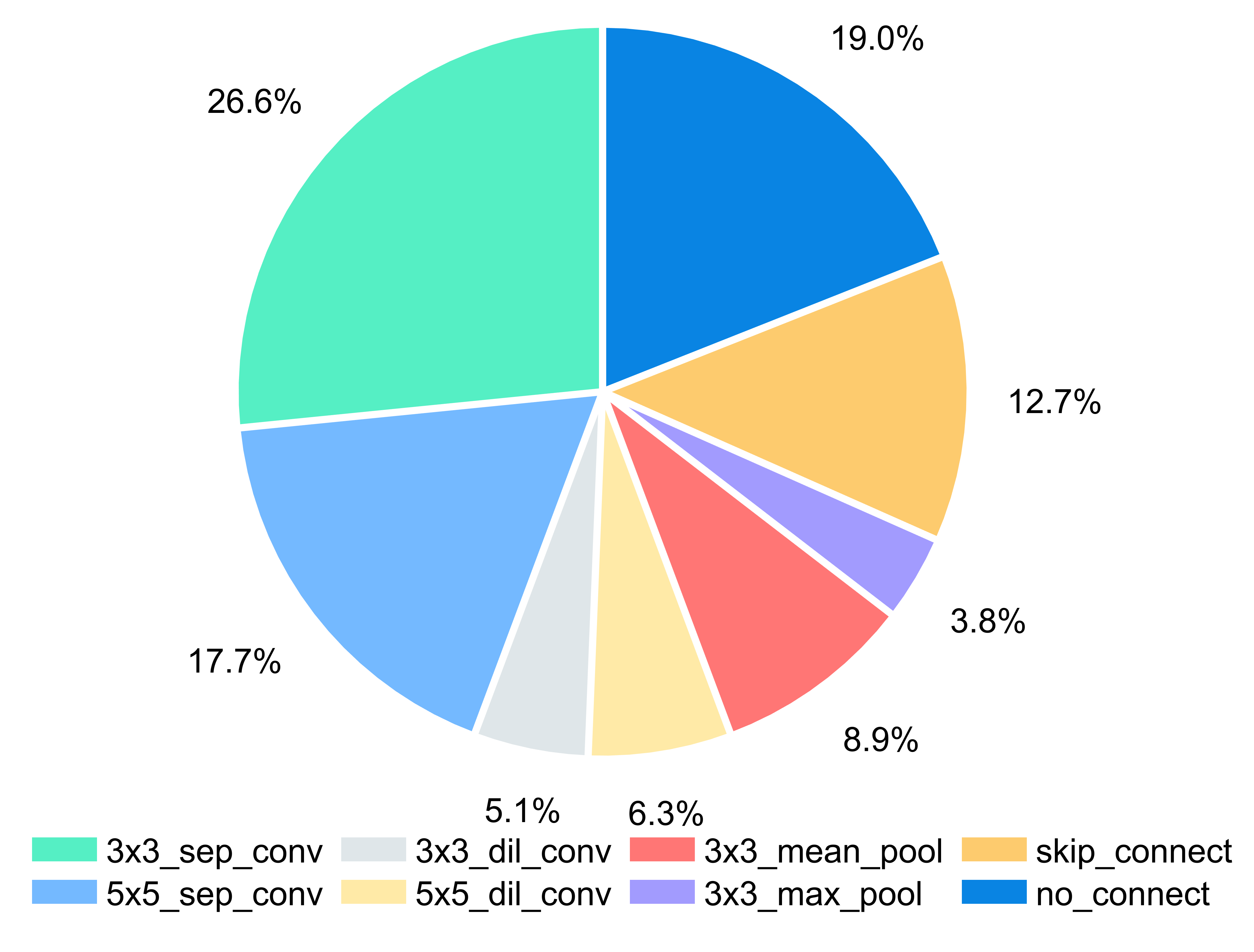}
    \caption{The proportion of each candidate operation in the cells of the encoder NAS of the final searched network on the CETUS dataset.}\label{encoder_3}
\end{figure}

\begin{figure}[ht]
    \centering
    \includegraphics[width=0.86\linewidth]{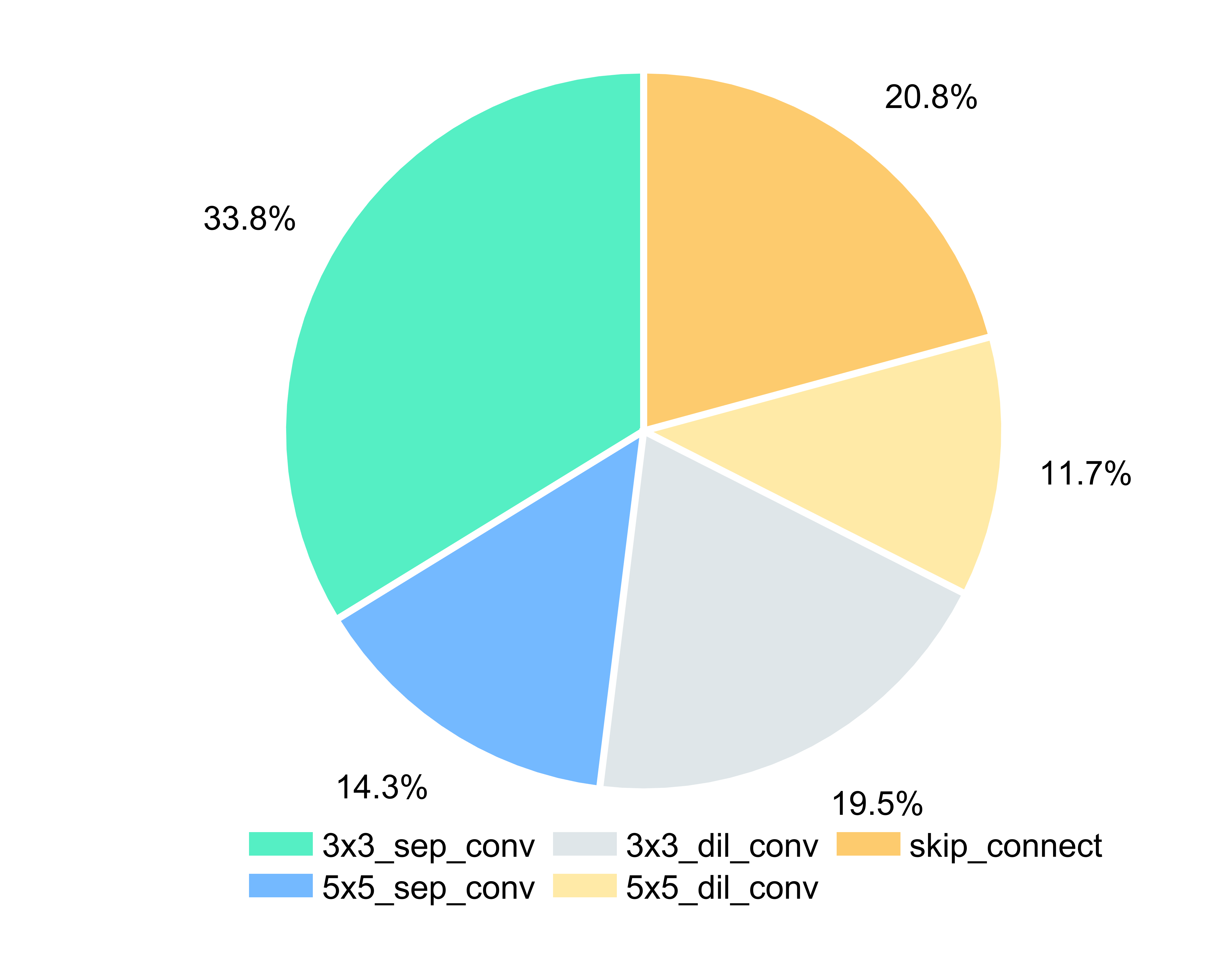}
    \caption{The proportion of each candidate operation in the cells of the decoder NAS of the final searched network on the CETUS dataset.}\label{decoder_3}
\end{figure}

As mentioned previously, we have defined that the cells within both the encoder and decoder architectures possess their distinct architecture parameters rather than sharing them. To gain a deeper insight into the final architecture, the statistical distributions of candidate operations in the final searched encoder and decoder architecture on three datasets (the CAMUS, the CETUS, and the HMC-QU) are presented in Fig. \ref{encoder_2} through Fig. \ref{decoder_3}, where the compositions of operations within the encoder and decoder cells are individually characterized.

The proportion of each candidate operation in the encoder and decoder cells is calculated as follows:

\begin{equation}
\begin{aligned}
    P_{O_{k}\in\mathcal{O}^{en}}=\frac{\sum_{i=1}^{\sum Cells}\sum_{j=1}^{\sum Edges} \alpha_{i,j,O_{k}}}{\sum Cells\sum Edges}
    \end{aligned}
\end{equation}
and
\begin{equation}
\begin{aligned}
    P_{O_{k}\in\mathcal{O}^{de}}=\frac{\sum_{i=1}^{\sum Cells} \alpha_{i,O_{k}}}{\sum Cells}
    \end{aligned}
\end{equation}
where $\sum Cells$ denotes the total number of the encoder or decoder cells, and $\sum Edges$ denotes the total number of node-to-node edges in the encoder. $\mathcal{O}^{en}$ and $\mathcal{O}^{de}$ represent the sets of candidate operations for the encoder and decoder, respectively.

By comparing the two charts (encoder and decoder) for each dataset, it can be observed that while the decoder favors skip connections and a more diverse set of convolution operations, the encoder places greater emphasis on separable convolutions and also includes pooling operations. Specifically, the decoder architecture searched by NAS has a much higher proportion of dilated convolutions compared to the encoder architecture. This distinction in operation distributions between the decoder and encoder components likely reflects their respective roles and computational requirements within the overall architecture, with the decoder emphasizing skip connections and diverse convolutions for upsampling and feature synthesis, while the encoder focuses on efficient feature extraction through separable convolutions and pooling operations.

\begin{table*}[t]
  \centering
  \begin{threeparttable}
  \begin{adjustbox}{width=\linewidth}
  \begin{tabular}{ccccc|ccc|ccc}
    \toprule
    \multirow{2}{*}{No.} & {Efficient} & \multirow{2}{*}{$\mathcal{L}_{uns}$} & \multirow{2}{*}{$\mathcal{L}_{ind}$} & \multirow{2}{*}{$\mathcal{L}_{con}$}  & \multicolumn{3}{c|}{CAMUS (50\%)} & \multicolumn{3}{c}{CETUS (50\%)} \\ 
    \cmidrule(){6-11}
    & NAS-ViT & & & &  DSC$\uparrow$ & IoU$\uparrow$ & 95HD$\downarrow$ & DSC$\uparrow$ & IoU$\uparrow$ & 95HD$\downarrow$ \\ 
    \midrule
    1 & \XSolidBrush & \XSolidBrush & \XSolidBrush & \XSolidBrush  & $0.918_{(0.003)}$ & $0.854_{(0.003)}$ & $6.587_{(0.440)}$ & $0.951_{(0.005)}$ & $0.967_{(0.004)}$ & $2.357_{(0.196)}$  \\ 
    2 & \Checkmark & \XSolidBrush & \XSolidBrush & \XSolidBrush  & $0.924_{(0.004)}$ & $0.860_{(0.007)}$ & $5.924_{(0.566)}$ & $0.959_{(0.002)}$ & $0.969_{(0.002)}$ & $2.049_{(0.099)}$  \\ 
    3 & \XSolidBrush & \Checkmark & \XSolidBrush & \XSolidBrush  & $0.923_{(0.002)}$ & $0.860_{(0.003)}$ & $6.024_{(0.313)}$ & $0.959_{(0.002)}$ & $0.970_{(0.001)}$ & $2.064_{(0.103)}$  \\ 
    4 & \XSolidBrush & \Checkmark & \Checkmark & \XSolidBrush  & $0.930_{(0.003)}$ & $0.871_{(0.004)}$ & $5.329_{(0.277)}$ & $0.962_{(0.001)}$ & $0.974_{(0.001)}$ & $1.926_{(0.051)}$ \\ 
    5 & \XSolidBrush & \Checkmark & \XSolidBrush & \Checkmark 
    & $0.927_{(0.005)}$ & $0.864_{(0.006)}$ & $5.509_{(0.501)}$& $0.960_{(0.003)}$ & $0.971_{(0.002)}$ & $2.011_{(0.148)}$  \\ 
    6 & \Checkmark & \Checkmark & \XSolidBrush & \XSolidBrush  
    & $0.929_{(0.002)}$ & $0.869_{(0.002)}$ & $5.395_{(0.264)}$& $0.963_{(0.002)}$ & $0.975_{(0.002)}$ & $1.903_{(0.115)}$ \\ 
    7 & \Checkmark & \Checkmark & \Checkmark & \XSolidBrush 
    & $0.934_{(0.001)}$ & $0.879_{(0.002)}$ & $5.205_{(0.189)}$ & $0.968_{(0.001)}$ & $0.977_{(0.001)}$ & $1.788_{(0.047)}$ \\
    \rowcolor[gray]{0.9}
    8 & \Checkmark & \Checkmark & \Checkmark & \Checkmark & {{$\boldsymbol{0.937_{(0.002)}}$}} & {{$\boldsymbol{0.884_{(0.003)}}$}} & {{$\boldsymbol{5.042_{(0.168)}}$}} & {{$\boldsymbol{0.972_{(0.001)}}$}} & {{$\boldsymbol{0.978_{(0.001)}}$}} & {{$\boldsymbol{1.620_{(0.036)}}$}}  \\
    \bottomrule  
\end{tabular}
    \end{adjustbox}
    \end{threeparttable}
    \caption{Ablation studies of each component in the proposed network structure on the HMC-QU, the CAMUS, and the CETUS datasets. }\label{ablation_submodules}
  \end{table*}

\section{More Ablation Studies}\label{app_sec:more ablation studies} 

\subsection{Effects of Sub-modules}\label{app_sec:effects of submodules}

To analyze the effects of each sub-module, additional experiments on the CAMUS and the CETUS datasets are conducted in {Table} \ref{ablation_submodules}. The baseline model (No. 1) is the NAS backbone (Sec \ref{NAS backbone}) without the Efficient NAS-ViT module, trained using 50\% of the annotated data. Firstly, incorporating the Efficient NAS-ViT module improves the segmentation performance from a DSC of 0.918 to 0.924 on the CAMUS dataset, consistent with the expectation of enhancing the context extraction capabilities. After introducing the co-training algorithm as the SSL framework (No. 6), DeNAS-ViT obtains 0.5\% and 0.4\% DSC improvements on the CAMUS and the CETUS datasets, respectively, attributed to the additional information gained from the unlabeled data facilitated by the SSL. Furthermore, when the network independence loss is considered (No. 7), there are 3.52\% and 6.04\% 95HD increments over the previous version (No. 6) on the two datasets respectively. The effectiveness of contrastive learning is validated in No. 5, which improves the performance by 0.4\% in DSC on the CAMUS dataset compared to independently using the $\mathcal{L}_{uns}$ in No. 3.

\subsection{Effect of Candidate Operation Set}\label{effect of candidate operation}

\begin{table}[ht]
  \centering
  \begin{threeparttable}
  \begin{adjustbox}{width=\linewidth} 
    \begin{tabular}{cccc}
    \toprule
    {Method} & DSC$\uparrow$ & IoU$\uparrow$ & 95HD$\downarrow$  \\  
    \midrule
    Include pooling and & \multirow{2}{*}{$0.924_{(0.005)}$} & \multirow{2}{*}{$0.923_{(0.004)}$} & \multirow{2}{*}{$2.961_{(0.238)}$} \\
    no connection &  & & \\
    \rowcolor[gray]{0.9}
    \textbf{Exclude pooling and} &  &  &  \\
    \rowcolor[gray]{0.9}
    \textbf{no connection} & \multirow{-2}{*}{$\boldsymbol{0.933_{(0.002)}}$}  & \multirow{-2}{*}{$\boldsymbol{0.931_{(0.002)}}$} & \multirow{-2}{*}{$\boldsymbol{2.480_{(0.161)}}$}\\
    \midrule
    \midrule
    Gain & $\uparrow 0.009$ & $\uparrow 0.008$ & $\downarrow 0.481$  \\
    \bottomrule
    \end{tabular} 
    \end{adjustbox}
    \end{threeparttable}
    \caption{Ablation study on the set of decoder candidate operations on the HMC-QU dataset.}\label{ablation_candidate}
  \end{table}
  
Table \ref{ablation_candidate} presents an ablation study that investigates the effect of including or excluding pooling and no connection operations from the set of candidate operations for the decoder component, evaluated on the HMC-QU dataset. The differences between the two scenarios are quantified in the final row, demonstrating that excluding pooling and no connection operations from the decoder cells leads to superior performance across all three evaluation metrics.

These empirical findings suggest that, for the given task and dataset, a simpler decoder architecture without pooling and no connection operations is more effective in achieving superior segmentation performance. The inclusion of these operations appears to be detrimental, potentially introducing noise or feature degradation that outweighs any potential benefits.

\begin{table}[t]
  \centering
  \begin{threeparttable}
    \begin{adjustbox}{width=0.99\linewidth}
    \begin{tabular}{c|ccc}
    \toprule
    {Method} & {DSC$\uparrow$} & {IoU$\uparrow$} & {95HD$\downarrow$} \\ 
    \midrule
    None & $0.924_{(0.002)}$ & $0.922_{(0.001)}$ & $2.852_{(0.102)}$ \\
    $r\in\{4,8\}$ & $0.929_{(0.001)}$ & $0.928_{(0.001)}$ & $2.603_{(0.115)}$  \\
    \rowcolor[gray]{0.9} 
    $\boldsymbol{r\in\{4,8,16\}}$ & $\boldsymbol{0.933_{(0.002)}}$ & $\boldsymbol{0.931_{(0.002)}}$ & $\boldsymbol{2.480_{(0.161)}}$  \\
    $r\in\{4,8,16,32\}$ & $0.927_{(0.002)}$ & $0.926_{(0.001)}$ & $2.719_{(0.148)}$  \\
    \bottomrule
    \end{tabular}
    \end{adjustbox}
    \end{threeparttable}
    \caption{Ablation Studies on different stages calculated in contrastive loss $\mathcal{L}_{con}$ on the HMC-QU dataset.}\label{ablation_4}
  \end{table}

\subsection{Impact of the Constraint at Different Feature Resolutions.}\label{impact of contrastive}
We further investigate whether incorporating the contrastive loss $\mathcal{L}_{con}$ at different feature resolutions can improve the model performance by gradually combining intermediate features from $r\in\{4,8,16,32\}$. As shown in {Table} \ref{ablation_4}, the performance is promoted by gradually incorporating resolutions from $r=4$ to $r=16$. Nevertheless, the performance decreases when leveraging the $r=32$. This is mainly because feature maps at $r=32$ contain high-level semantic information but lack detailed information necessary for contrastive analysis, and then the loss computed at this resolution does not contribute to the improvement.

\section{More Hyper-parameter Studies}\label{hyper-parameter study}

\begin{figure}[t]
\centering
    \begin{subfigure}{0.49\linewidth}
        \centering
        \includegraphics[width=\linewidth]{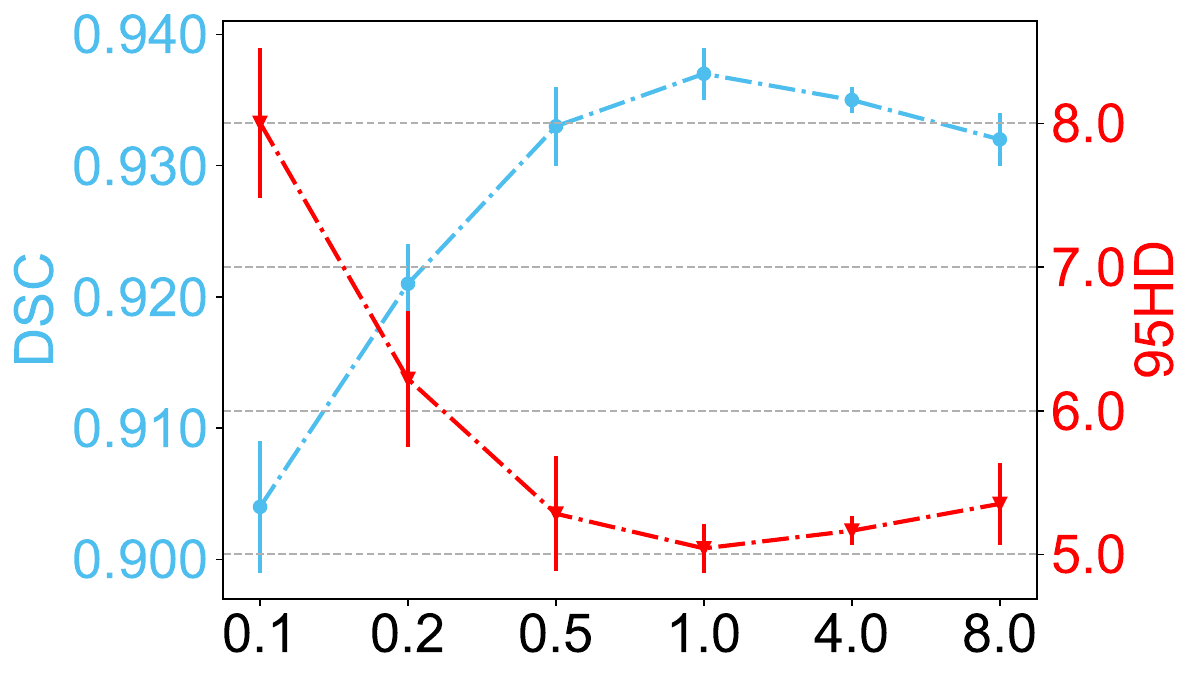}
        \caption{Supervised constraint\\coefficient $\lambda_{1}$.}
    \end{subfigure}
    \begin{subfigure}{0.49\linewidth}
        \centering
        \includegraphics[width=\linewidth]{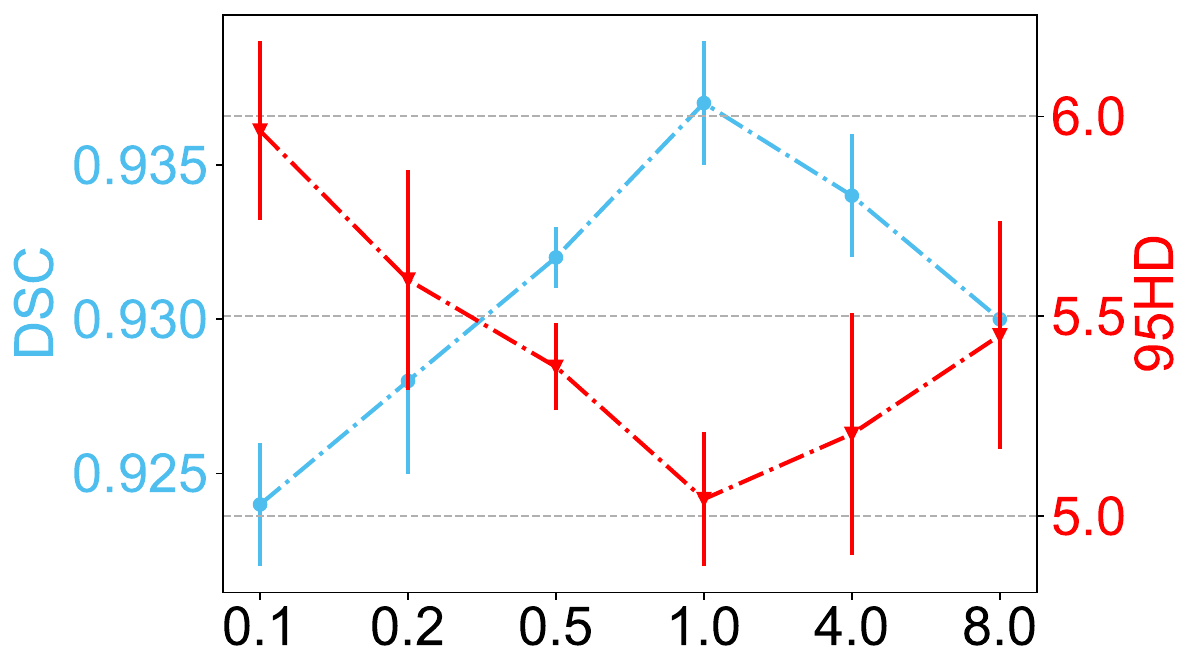}
        \caption{Unsupervised and contrastive constraint coefficients $\lambda_{2},\lambda_{4}$.}
    \end{subfigure}
    \begin{subfigure}{0.49\linewidth}
        \centering
        \includegraphics[width=\linewidth]{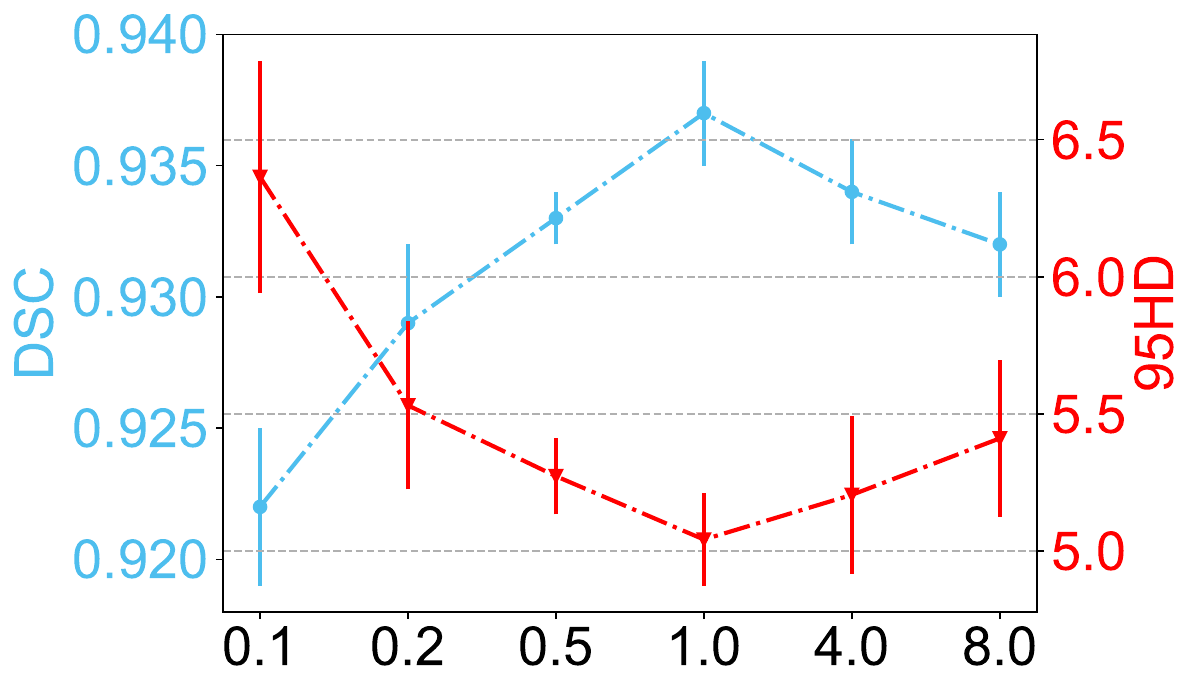}
        \caption{Independence constraint coefficient $\lambda_{3}$.}
    \end{subfigure}
    \begin{subfigure}{0.49\linewidth}
        \centering
        \includegraphics[width=\linewidth]{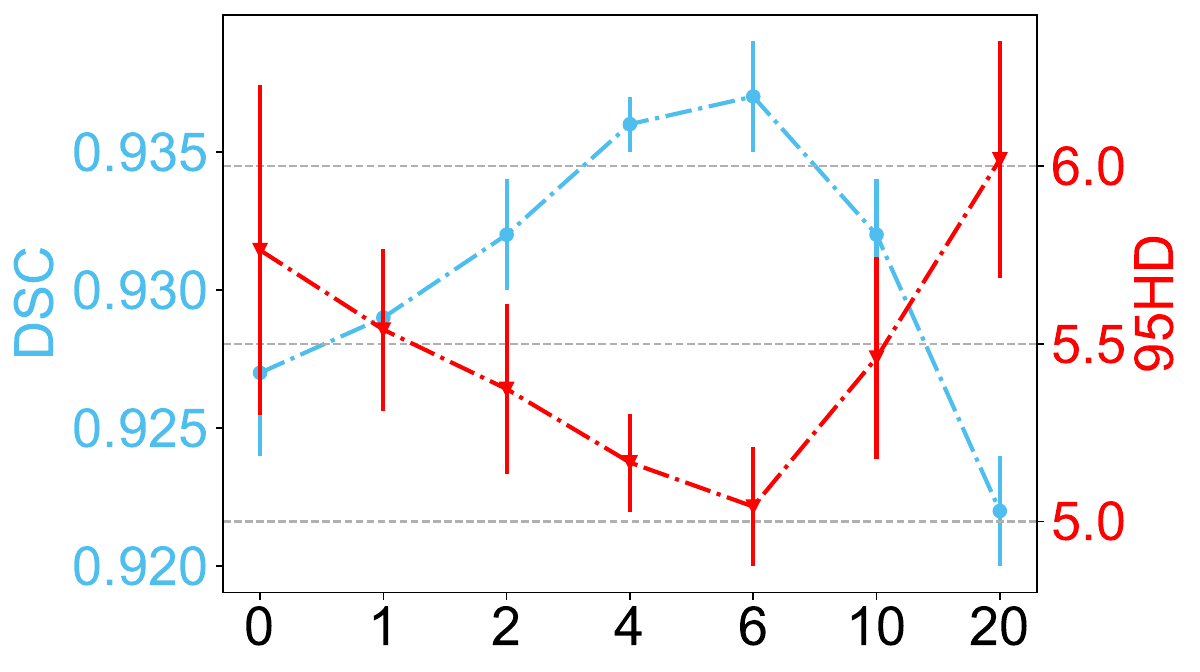}
        \caption{Independence constraint optimization epochs $E_{B}$}
    \end{subfigure}
    \caption{Hyper-parameter study on constraints. The balance between constraints has an impact, but it is not significant.}\label{ablation_3}
\end{figure} 

\subsection{Sensitivity to Loss Coefficients.}\label{app_sec:sensitivity to loss}

The sensitivity to hyper-parameter settings of loss coefficients is further discussed. As shown in {Fig.~\ref{ablation_3}(a)}, the fluctuations in DSC caused by varying $\lambda_{1}$ indicate that the supervised loss plays a dominant role. From {Fig.~\ref{ablation_3}(b)}, we observe that assigning a low weight to the initial value of $\lambda_{2}$ weakens the role of the co-training. While setting to a high value, it causes DeNAS-ViT to deviate from the main segmentation task. This principle is also applicable to the contrastive loss coefficient $\lambda_{3}$. As observed in {Fig.~\ref{ablation_3}(c)(d), the degree of emphasis on network independence has an obvious influence. When $\lambda_{4}=2$ and $E_{B}=6$, we obtain the highest DSC, whereas assigning a small weight to the independence loss fails to utilize the advantage of co-training, and a small number of iterations also lead to insufficient network independence estimation. On the other hand, an excessive focus on independence causes the NAS backbone to deviate from the optimum.

\subsection{Sensitivity to Network Layers}\label{app_sec:sensitivity to layers}

\begin{table}[ht]
  \centering
  \begin{adjustbox}{width=\linewidth}
  \begin{threeparttable}
    \begin{tabular}{c|ccc}
    \toprule
    {Method} & {HMC-QU} & {CAMUS} & {CETUS} \\ 
    \midrule
    $4$ & $0.920_{(0.002)}$ & $0.866_{(0.002)}$ & $0.969_{(0.002)}$  \\
    $6$ & $0.928_{(0.001)}$ & $0.879_{(0.003)}$ & \cellcolor[gray]{0.9}$\boldsymbol{0.978_{(0.002)}}$  \\
    \rowcolor[gray]{0.9}
     $\boldsymbol{8}$ & \cellcolor[gray]{0.9}$\boldsymbol{0.931_{(0.002)}}$ & \cellcolor[gray]{0.9}$\boldsymbol{0.884_{(0.003)}}$ & \cellcolor[gray]{0.9}$\boldsymbol{0.978_{(0.001)}}$  \\
    \bottomrule
    \end{tabular} 
    \end{threeparttable}
    \end{adjustbox}
    \caption{Hyper-parameter study on the impact of different numbers of network layers $L$ on three datasets. The evaluation metric is IoU.}\label{hyper-parameter}
  \end{table}

To evaluate the effect of the number of network layers $L$, the network is initialized with 4, 6, and 8 layers, respectively, while maintaining the other parameters constant to facilitate a fair comparison. The quantitative results obtained across the HMC-QU, the CETUS, and the CAMUS datasets are presented in Table \ref{hyper-parameter}. It can be observed that the proposed model employing 8 layers achieves superior performance in comparison to using 4 and 6 layers. Based on this finding, $L$ in our model is initialized to 8. 

\section{Limitations and Future Work}\label{app_sec:limitations}
Our future work is to further enhance the zero-shot generalization capability (zero-shot learning) of our model. Specifically, while our current approach involves training on ultrasound image datasets, we aim to investigate the model's transferability and robustness by evaluating its performance on other medical imaging modalities, such as magnetic resonance imaging (MRI) or computed tomography (CT) scans. In other words, we hope that the model can effectively perform tasks without having encountered specific examples of those tasks or domains during training.

This cross-modality research will provide insights into the model's ability to generalize and adapt to diverse imaging data, which is crucial for practical clinical applications. By enhancing zero-shot generalization capability across modalities, we strive to develop a more versatile and robust solution for medical image analysis workflows, ultimately contributing to improved patient care.

\end{document}